\pdfoutput=1

\documentclass{article}
\PassOptionsToPackage{table}{xcolor}  %
\usepackage{arydshln}  %

\usepackage{iftex}
\ifPDFTeX
  \usepackage{iclr2027_conference,times}
\else
  \usepackage{iclr2027_conference}
  \usepackage{fontspec}
\fi

\usepackage{amsmath,amsfonts,bm}

\def\eqref#1{equation~\ref{#1}}

\def\1{\bm{1}}

\DeclareMathAlphabet{\mathsfit}{\encodingdefault}{\sfdefault}{m}{sl}
\SetMathAlphabet{\mathsfit}{bold}{\encodingdefault}{\sfdefault}{bx}{n}

\usepackage{hyperref}
\usepackage{url}
\usepackage{amsmath,amssymb}
\usepackage{graphicx}
\usepackage{float}                %
\usepackage{wrapfig}              %
\usepackage{needspace}            %
\usepackage{pgfplots}             %
\pgfplotsset{compat=1.17}
\usetikzlibrary{calc}
\usepackage{booktabs}

\newcommand{\tabstyle}{\footnotesize\setlength{\tabcolsep}{3pt}\renewcommand{\arraystretch}{1.1}}
\newcommand{\tabstyledense}{\scriptsize\setlength{\tabcolsep}{3pt}\renewcommand{\arraystretch}{1.1}}
\usepackage{wrapfig}              %

\usepackage{multirow}
\usepackage{xcolor}               %
\usepackage{arydshln}             %
\usepackage{subcaption}
\usepackage[capitalize]{cleveref}
\usepackage{etoc}                 %

\newcommand{\method}{ReDiverse}          %
\newcommand{\methodft}{\method{}-FT}      %
\newcommand{\methodtf}{\method{}-TF}      %
\newcommand{\ours}{\method{}}              %
\newcommand{\ourstrain}{\methodft{}}       %
\newcommand{\ourstest}{\methodtf{}}        %

\newlength{\steerimgw}   %

\usepackage{mdframed}
\usepackage{enumitem}   %
\newmdenv[
  linewidth=0.9pt,
  linecolor=black!55,
  backgroundcolor=black!4,
  innertopmargin=7pt, innerbottommargin=7pt,
  innerleftmargin=9pt, innerrightmargin=9pt,
  skipabove=10pt, skipbelow=6pt
]{findingbox}

\newcommand{\qedsq}{\ifmmode\tag*{$\square$}\else\hfill$\square$\fi}

\definecolor{groupbg}{HTML}{EDF1F6}   %
\definecolor{keycol}{HTML}{FFF4E0}    %
\definecolor{headfg}{HTML}{4A5560}
   \usepackage{booktabs,multirow,graphicx,amssymb}
   \usepackage[table]{xcolor}          %

\usepackage{booktabs}
\usepackage{xcolor}

\definecolor{BlockGray}{gray}{0.88}   %
\definecolor{TextGray}{gray}{0.50}    %
\definecolor{OursGreen}{RGB}{225,240,225}

\providecommand{\tg}[1]{\textcolor{TextGray}{#1}}
\definecolor{AblBand}{RGB}{209,218,232}   %
\definecolor{AblPick}{RGB}{250,235,208}   %
\usepackage{tabularx}
\usepackage{multirow}
\usepackage{subcaption}

\newsavebox{\qrowbox}

\newsavebox{\qstripbox}

\usepackage{adjustbox}

\title{Restoring Diversity in Distilled Image and Video Generators by Probing Activations}
\usepackage{fontawesome5}
\newcommand{\urlhomepage}{https://rediverse-distillation.github.io/rediverse-distillation/}  %
\newcommand{\urlgithub}{https://github.com/H-furkan/rediverse}       %
\newcommand{\urlhf}{https://huggingface.co/ReDiverse/ReDiverse}          %
\definecolor{projlink}{HTML}{3B6EA8}
\newcommand{\projlinks}{{\hypersetup{hidelinks}\normalsize%
  \href{\urlhomepage}{\faHome~\textcolor{projlink}{Homepage}}\qquad
  \href{\urlgithub}{\faGithub~\textcolor{projlink}{GitHub}}\qquad
  \href{\urlhf}{\raisebox{-0.22ex}{\includegraphics[height=1.1em]{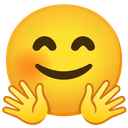}}~\textcolor{projlink}{Hugging Face}}}}
\author{
Furkan Horoz$^{1}$\quad
Ozgur Kara$^{1}$\quad
Kaan Oktay$^{2}$\quad
Adil Kaan Akan$^{2}$\\
\textbf{Atakan Kara$^{2}$\quad
Anand Bhattad$^{3\dagger}$\quad
James Matthew Rehg$^{1\dagger}$}\\[4pt]
$^{1}$University of Illinois at Urbana-Champaign\quad
$^{2}$fal\quad
$^{3}$Johns Hopkins University\\[6pt]
\projlinks
}
\iclrfinalcopy
\makeatletter
\def\@maketitle{\vbox{\hsize\textwidth\centering
{\LARGE\sc \@title\par}
\lhead{}
\begin{tabular}[t]{c}\bf\rule{\z@}{20pt}\@author\end{tabular}\par
\vskip 0.20in}}
\makeatother
\renewcommand{\headrulewidth}{0pt}   %

\begin{document}
\maketitle
{\renewcommand{\thefootnote}{}\footnotetext{$^\dagger$Equal advising.}}
\etocdepthtag.toc{main}

\begin{figure}[H]
  \vspace{-15pt}
  \centering
  \includegraphics[width=0.94\linewidth]{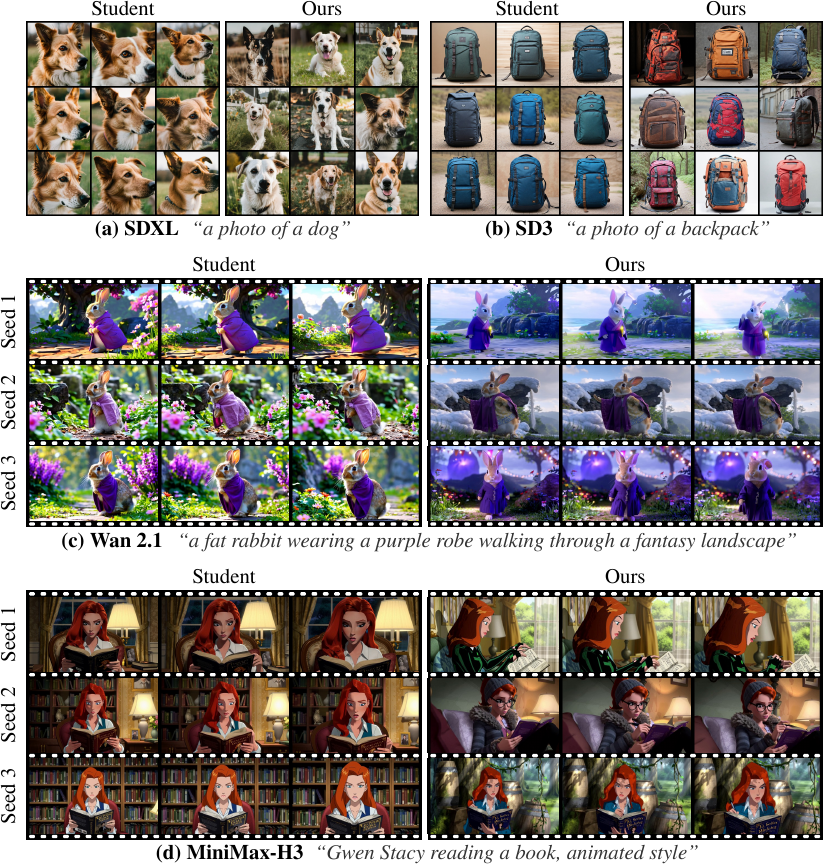}
\caption{\looseness=-1 \textbf{ReDiverse improves diversity in distilled image and video generators.}  For a fixed prompt, the student tends to repeat similar visual compositions, while ReDiverse produces broader variation: (a) dogs appear at different viewing distances and poses; (b) backpacks in various styles; and (c--d) videos show diverse environments, viewpoints, and styles. Each image grid shows samples from different seeds; each video row corresponds to one seed, with time progressing from left to right.}
  \label{fig:teaser}
\end{figure}

\begin{abstract}
Distilling diffusion models into few-step student models substantially reduces sampling latency, but induces severe diversity collapse: samples generated from different random seeds show minimal variation. 
In this work, we investigate the root cause of this failure by probing the internal representations of student models relative to their teachers. 
We find that distilled students maintain total activation variance comparable to their teachers, but concentrate it into substantially fewer directions. 
This directional collapse is localized to an identifiable intermediate block at the initial denoising step. 
Consequently, diversity collapse stems from a misallocation of activation variance rather than an inherent loss of generative capacity. 
Guided by this finding, we propose \method{} to reinstate the missing variance directions. 
At training time, \methodft{} fine-tunes a lightweight adapter that regresses student activations onto the teacher's, reviving the teacher's feature variance characteristic. 
At inference time, \methodtf{} introduces a training-free mechanism that steers activations along precomputed teacher principal directions. 
Across image and video distillation benchmarks, \method{} consistently restores sample diversity while preserving generation quality, further supported by the user study.
\vspace{-8pt}
\end{abstract}

\section{Introduction}
\label{sec:introduction}

Distillation turns diffusion-based image and video generators~\citep{ho2020denoising,song2021scorebased,rombach2022high} into fast, few-step models~\citep{salimans2022progressive,pmlr-v202-song23a,yin2024one,yin2024improved}. But they can reduce the diversity of their outputs: different seeds may produce strikingly similar samples for the same prompt. We explore where in the student diversity disappears. 

Broadly, there are two solutions to alleviate this problem. (1) Retrain or re-distill the student, either with a modified or regularized objective \citep{xu2025one,lu2025adversarial,wu2026diversitypreserved,li2026duet} or with real data \citep{chen2026dataforcing}. (2) Keep the student frozen and intervene at inference, either by optimizing the initial noise \citep{harrington2026s}, by perturbing intermediate features directly in the forward pass \citep{yadav2026stride,bhat2026don}, or by running the teacher for the first step \citep{gandikota2026distilling}.  These approaches motivate a complementary question: where does the student's internal representation cease to preserve the variation present in its teacher, and can that diagnosis guide a targeted repair?

We start by asking a simple question. What would happen to the activations across many seeds when the same prompt is given to a teacher and its student? Surprisingly, we found a similar pattern emerging in every model we tested, across both image and video families. In particular, at each block, we measure the total activation variance across seeds and its participation ratio, which summarizes how broadly that variance is distributed across directions.
We find that in the initial blocks, the student's total variance and participation ratio closely match the teacher's. However, deeper down, the student's total variance equals or exceeds the teacher's, but it is restricted to far fewer directions in a particular block, resulting in a cascaded effect on downstream activations. Interestingly, we show that the collapse in diversity can be attributed to these lost directions and misallocation of variance within a single block of the student's model. 

\looseness=-1 This diagnosis identifies where to supervise the student to recover a richer representation. We introduce \textbf{\method{}} with two complementary interventions. In \textbf{\methodft{}}, we insert LoRA modules into all layers up to and including the identified block, keeping the base model frozen. We train these modules by regressing the student's activation at this block during its first denoising step toward the teacher's activation at the corresponding block during its $\tau$-th denoising step. At inference, the LoRA modules are active only during the student's first step; subsequent steps use the original student. In \textbf{\methodtf{}}, no training is required: we perturb activations at the identified block along teacher principal directions computed once per prompt, with random coefficients scaled by the teacher's spread along each direction. The two interventions can be used independently or combined.

To summarize, our contributions are threefold:
\begin{itemize}[leftmargin=1.4em,itemsep=2pt,topsep=2pt]
\item We profile and probe the variance characteristics of distilled students against their teachers, block by block under identical inputs, and show that distillation keeps the total variance but concentrates it into fewer directions at an identifiable block. This locates the collapse, which output metrics cannot, and separates students distilled from the same teacher under different recipes.
\item We introduce \method{} to restore variation at the identified block through two complementary interventions: \methodft{} trains LoRA modules in the layers up to and including the identified block to align first-step student activations with later-step teacher activations, while \methodtf{} perturbs activations along teacher principal directions without training.
\item We evaluate both interventions and their combination on four teacher--student pairs spanning image (SDXL~\citep{podell2024sdxl}, SD3~\citep{esser2024scaling}) and video (Wan~2.1~\citep{wan2025wan}, MiniMax-H3~\citep{minimax2026h3}) generators, where \method{} restores sample diversity while preserving generation quality, further supported by a user study.
\end{itemize}

\section{Related Work}
\label{sec:related_work}

\paragraph{Distillation and Diversity.}
Distillation compresses iterative diffusion models \citep{ho2020denoising,song2021scorebased,rombach2022high,podell2024sdxl} into few-step students by matching teacher trajectories \citep{salimans2022progressive}, consistency endpoints \citep{pmlr-v202-song23a,luo2024latent}, or output distributions via adversarial \citep{sauer2024adversarial} and distribution-matching (DMD) losses \citep{wang2023prolificdreamer,yin2024one,yin2024improved,luo2025learning}, recently extended to video transformers \citep{huang2026self,zheng2026large,zheng2026causal,shaul2026parallel,fastvideo2026,lightx2v2026turbo}. Because these objectives supervise terminal outputs rather than seed-conditioned activations, distilled students prematurely freeze global structure at the first step \citep{gandikota2026distilling} and generate near-identical outputs across seeds \citep{shi2026manifold,li2026duet}. This collapse has been attributed to mode-seeking reverse KL \citep{minka2005divergence,xu2025one,lu2025adversarial}, adversarial training \citep{arora2017generalization}, constrained step budgets \citep{salmona2022can,fan2026phased}, or over-guided high-noise targets \citep{li20261}. Existing remedies either retrain the student with modified divergences \citep{xu2025one,lu2025adversarial}, consistency--DMD hybrids \citep{zheng2026large,li2026duet}, adjusted first-step targets \citep{wu2026diversitypreserved,li20261}, or real data \citep{chen2026dataforcing}, incurring substantial re-distillation costs and fidelity trade-offs \citep{li2026duet}; or intervene at inference on a frozen student via noise optimization \citep{harrington2026s,shi2026manifold}, teacher--student hybrid sampling \citep{gandikota2026distilling}, or feature perturbations \citep{bhat2026don,yadav2026stride}. Crucially, neither paradigm localizes where diversity vanishes internally. \method{} identifies this breakdown and is agnostic to the distillation recipe.

\vspace{-8pt}
\paragraph{Interpreting diffusion representations.}
Prior work has shown that the intermediate activations of a diffusion model encode attributes of the samples it generates: the U-Net bottleneck forms a semantic space \citep{kwon2023diffusion}, whose editing directions can be found from principal components of these features \citep{haas2024discovering}, the encoder's pullback metric \citep{park2023understanding}, or the low-rank Jacobian of the posterior mean \citep{chen2024exploring}; all study a single model. In distilled students, \citet{gandikota2026distilling} find that the directions needed for diversity survive but go unused, and \citet{yadav2026stride} inject noise along the student's own principal directions, spreading variance it still has. However, we are the first to interpret student and teacher activations together, both to explain why diversity collapses and to remedy it.
\section{Probing Distilled Diffusion Models}
\label{sec:diagnosis}

To identify the internal mechanisms driving diversity collapse, we analyze the representations of distilled diffusion models across network depth and denoising steps. Specifically, we trace the propagation of seed sensitivity by comparing intermediate activations between teacher models and their few-step student counterparts under identical initial latents and conditioning prompts.

\vspace{-5pt}
\subsection{Same Variance, Fewer Directions}
\label{sec:diagnosis_collapse}

To characterize the internal representation of block $b$, we evaluate three geometric quantities derived from the sample covariance matrix $\Sigma_b$ across random noise initializations: total variance (TV) $\mathrm{Tr}(\Sigma_b)$, measuring overall dispersion; the participation ratio $\mathrm{PR}(\Sigma_b)$ \citep{PPR:PPR23693}, quantifying the effective dimensionality spanned by this variance, as eigenvalue-spectrum measures of dimensional collapse do in representation learning \citep{jing2022understanding,garrido2023rankme}; and the concentration ratio $\mathrm{Tr}(\Sigma_b) / \mathrm{PR}(\Sigma_b)$, measuring the average variance allocated per active dimension. 

\begin{figure}[t]
  \centering
  \includegraphics[width=\linewidth]{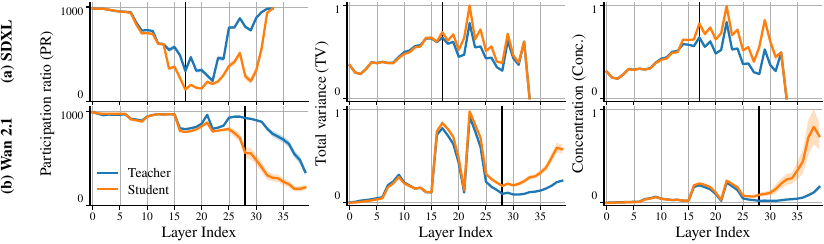}
  \caption{\textbf{Variance characteristic along the depth.} Participation ratio, total variance and concentration of
    teacher and student activations at every layer, for (a) an image model (SDXL) and (b) a video
    model (Wan~2.1). Total variance and concentration are min--max normalized to $[0,1]$ over teacher and student
    (SDXL's in log space). The solid vertical line marks the collapsed block.}
  \label{fig:pr_depth}
\end{figure}

We find that diversity collapse is driven by a directional bottleneck rather than a loss of variance: distilled students preserve total activation variance, but confine it to fewer directions starting at an intermediate layer (Fig.~\ref{fig:pr_depth}). Across all depths, the student's total variance $\mathrm{Tr}(\Sigma_b)$ matches or exceeds that of the teacher at most blocks, confirming that overall variance is not depleted. While early layers track the teacher, the participation ratio $\mathrm{PR}(\Sigma_b)$ drops at an intermediate layer---the \emph{collapsed block}.   \Needspace*{14\baselineskip}
\begin{wraptable}{r}{0.40\linewidth}
  \vspace{-\intextsep}
  \centering
  \caption{\textbf{Student/teacher ratios.} The student's value divided by the teacher's at the collapsed
  block.}
  \label{tab:pr_summary}
  \tabstyle
  \begin{tabularx}{\linewidth}{@{}l >{\centering\arraybackslash}X >{\centering\arraybackslash}X >{\centering\arraybackslash}X@{}}
    \toprule
    Student & PR & TV & Conc. \\
    \midrule
    DMD2 (SDXL)     & 0.40 & 1.63 & 4.76 \\
    TDM (SD3)       & 0.49 & 1.03 & 2.68 \\
    lightx2v (Wan)  & 0.60 & 1.41 & 2.54 \\
    Turbo (H3)      & 0.37 & 2.92 & 8.08 \\
    \bottomrule
  \end{tabularx}
  \vspace{-0.5\intextsep}
\end{wraptable}

Tab.~\ref{tab:pr_summary} divides each student's value at the collapsed block by its teacher's, showing how much the student amplifies or shrinks each quantity, for every model. Across all evaluated architectures and modalities, students preserve or even amplify total variance ($1.03\times\text{--}2.92\times$) while retaining merely $37\%\text{--}60\%$ of the teacher's directional span. Consequently, the surviving axes are overloaded with up to $8\times$ the variance observed in the teacher, driving a sharp surge in concentration that propagates through all downstream layers to the final output. While the exact layer index depends on network topology, this directional compression occurs universally, identifying it as an intrinsic hallmark of few-step distillation.  

\vspace{-5pt}
\subsection{Fewer Directions, Less Diversity}\label{sec:diagnosis_steering}
\Needspace*{18\baselineskip}   %
\begin{wrapfigure}{r}{0.42\textwidth}
  \vspace{-28pt}
  \centering
  \includegraphics[width=\linewidth]{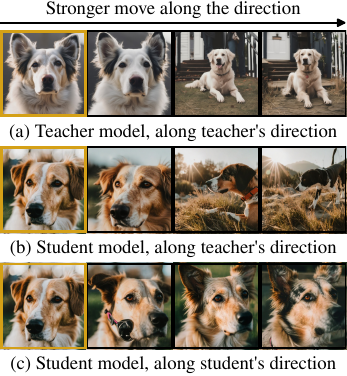}
  \caption{\textbf{Moving along a direction, in the teacher and in the student model.}}
  \label{fig:steering}
  \vspace{-10pt}
\end{wrapfigure}

Because downstream layers deterministically map intermediate features to the final output, the directional span of activation variance bounds attainable generative diversity. To test whether the directions lost during distillation govern macroscopic diversity and whether the student can still decode them, we perform activation steering at the collapsed block during the initial denoising step. Specifically, we extract the top principal direction of activation covariance across random noise initializations and perturb intermediate activations along it with increasing strength $\alpha$, scaled by the empirical standard deviation.

Perturbing the teacher along its own principal axis dynamically alters macroscopic attributes, including camera framing, subject pose, and scene composition (Fig.~\ref{fig:steering}a). Crucially, injecting this same teacher direction into the student produces matching structural changes (Fig.~\ref{fig:steering}b), recovering diverse poses and layouts without modifying any student weights. In contrast, steering the student along its own leading axis alters only appearance-level details, such as coat pattern and localized lighting, leaving global geometry invariant (Fig.~\ref{fig:steering}c).
This functional asymmetry confirms two key insights: (i) the directions suppressed during distillation are precisely those that dictate structural diversity, and (ii) downstream student layers retain the capacity to decode these modes once restored. Diversity collapse is therefore not an irreversible loss of generative capacity, but a geometric misallocation of activation variance. 

\Needspace*{16\baselineskip}   %
\begin{wrapfigure}[11]{r}{0.40\textwidth}
\vspace{-5pt}
  \centering
  \includegraphics[width=\linewidth]{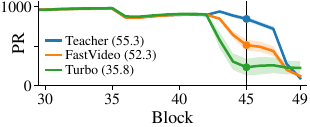}
  \caption{\textbf{PR along depth.} MiniMax-H3 and two students (DINO diversity score in parentheses); solid line:
  collapsed block.}
  \label{fig:pr_same_teacher}
  \vspace{-10pt}
\end{wrapfigure}

In addition, we found that two students distilled from the same teacher (MiniMax-H3 \citep{minimax2026h3}) reach different output diversity, and that their participation ratios at the collapsed block differ in the same order (Fig.~\ref{fig:pr_same_teacher}). Both track the teacher until just before the collapsed block (block 45), where FastVideo \citep{fastvideo2026} keeps 61\% of the teacher's participation ratio and Turbo \citep{lightx2v2026turbo} only 28\%, and FastVideo is correspondingly more diverse (DINO 52.3 vs.\ 35.8; teacher 55.3). The retained participation ratio at the collapsed block thus tracks output diversity.
\section{\method{}: Restoring the Lost Directions}
\label{sec:method}

We introduce \method{}, which operationalizes our diagnostic findings through two complementary mechanisms: (i) \method{}-FT (Sec.~\ref{sec:method_training}), a targeted fine-tuning framework that regresses student activations onto the teacher's at the collapsed block to revive the teacher's variance characteristic, and (ii) \method{}-TF (Sec.~\ref{sec:method_tf}), a training-free inference approach that injects variance along precomputed teacher principal directions to restore the directions the student lost; both hence improve diversity by adjusting the student's feature variance characteristic.

\vspace{-5pt}
\subsection{\methodft{}: Directional Alignment via Targeted Adaptation}
\label{sec:method_training}

\paragraph{Feature regression.}
To align the student's feature variance characteristic with the teacher's, \methodft{} regresses the student's activation at the collapsed block $\ell$ onto that of the frozen teacher, a FitNets-style feature regression \citep{romero2015fitnets} at one block and step. Let $h_{\ell,t}$ denote the activation at block $\ell$ during denoising step $t$. Both models start from the same initial noise $\epsilon$ and prompt $c$: the student is read at its first step ($t{=}1$), where it forms the layout of the sample from pure noise, and the teacher is run for $\tau{-}1$ steps and read at step $\tau$:
\begin{equation}
  \mathcal{L}_{\mathrm{reg}}(\theta)
  \;=\; \mathbb{E}_{\epsilon,\,c}\!\left[\frac{1}{D}
    \bigl\lVert h^{\theta}_{\ell,1}(\epsilon, c)
      - h^{\phi}_{\ell,\tau}(x^{\phi}_{\tau}, c) \bigr\rVert_2^2
  \right],
  \quad \text{where } x^{\phi}_{\tau} = \mathcal{S}^{\phi}_{\tau-1}(\epsilon, c).
  \label{eq:anchor}
\end{equation}
Here, $\mathcal{S}^{\phi}_{\tau-1}$ runs $\tau{-}1$ teacher steps from $\epsilon$, and $D$ is the dimension of the activation, so the loss is the mean squared error over its elements. Each prompt is paired with $M$ noises, so matching the activations noise by noise matches the teacher's variation across noises rather than a single activation. This aligns the orientation and relative energy of the feature manifold, natively restoring the suppressed eigenspectrum without requiring auxiliary covariance regularizers or adversarial objectives. The teacher step $\tau$, the teacher's guidance \citep{ho2021classifierfree} up to that step, the training prompts and $M$ are design choices (Apps.~\ref{app:step} and \ref{app:cfg}, Sec.~\ref{sec:experiments_settings}); by default the teacher is guided up to step $\tau$ and read in its conditional branch at step $\tau$.

\begin{figure}[t]
  \centering
  \includegraphics{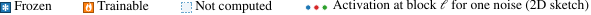}\\[3pt]   %
  \begin{subfigure}[t]{0.485\linewidth}
    \centering
    \includegraphics[width=\linewidth]{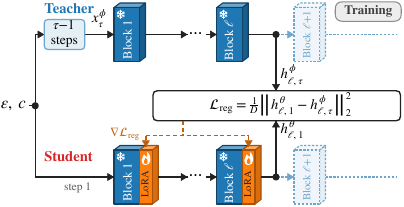}\\[3pt]
    \includegraphics[width=\linewidth]{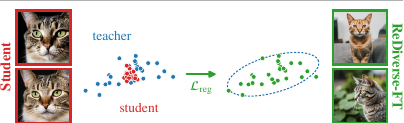}
    \caption{\methodft{} (``a photo of a cat'')}
    \label{fig:method_training}
  \end{subfigure}\hfill{\color{black!30}\vrule width 0.5pt}\hfill   %
  \begin{subfigure}[t]{0.485\linewidth}
    \centering
    \includegraphics[width=\linewidth]{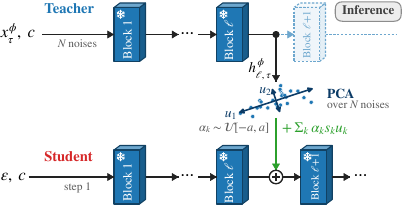}\\[3pt]
    \includegraphics[width=\linewidth]{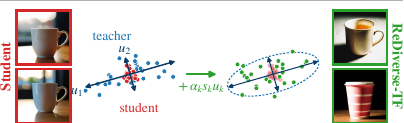}
    \caption{\methodtf{} (``a photo of a cup'')}
    \label{fig:method_inference}
  \end{subfigure}
  \caption{\textbf{\method{} overview.} The student loses the teacher's variation across noises at the collapsed block $\ell$, while the later blocks can still render it; \method{} restores it there, by training (a) or at inference (b). Below each diagram, the activations (blue: teacher, red: student, green: \method{}) and samples, before and after. (a)~\methodft{} regresses the student's activation onto the teacher's for the same noise, updating only LoRA adapters. (b)~\methodtf{} finds the teacher's directions of variation by PCA and moves the student along them.}
  \label{fig:method}
\end{figure}

\vspace{-8pt}
\paragraph{Localized LoRA adaptation.}
We implement \methodft{} using LoRA \citep{hu2022lora} inserted only into the blocks up to and including the collapsed block $\ell$: as these blocks map the initial noise into intermediate representations, adapting only this region enables the student to reproduce the teacher's variance characteristic before activations propagate downstream. Downstream layers already retain the capacity to decode diverse structural modes (Sec.~\ref{sec:diagnosis_steering}); leaving them frozen preserves native synthesis fidelity without disrupting the rendering pipeline. Because restoring the span redistributes variance across suppressed axes rather than learning new concepts, low-rank adaptation provides an inductive bias that recovers diversity without overfitting \citep{gan2026neural}.

\vspace{-5pt}
\subsection{\methodtf{}: Training-Free Stochastic Steering}
\label{sec:method_tf}

{\setlength{\abovedisplayskip}{3pt}\setlength{\abovedisplayshortskip}{0pt}%
\setlength{\belowdisplayskip}{3pt}\setlength{\belowdisplayshortskip}{3pt}%
Because downstream layers can decode the teacher's principal directions without parameter updates (Sec.~\ref{sec:diagnosis_steering}), we can restore diversity directly at test time by steering intermediate activations along these missing axes. To identify the exact directions to inject and calibrate their energy so that activations remain in distribution, we extract this directional subspace with principal component analysis \citep{bishop2006pattern} on the teacher, once per prompt before sampling. Specifically, for a given prompt, we run the teacher for $\tau{-}1$ steps across $N$ random noise initializations and record its centered activations at the collapsed block $\ell$ during step $\tau$, matching the regression target of Eq.~(\ref{eq:anchor}). Arranging these activations into a matrix $X \in \mathbb{R}^{N \times D}$, the eigendecomposition of the sample Gram matrix $X X^{\top}$ yields eigenvalues $\lambda_k$ and eigenvectors $v_k$. The $k$-th principal direction $u_k$ and the teacher's empirical spread $s_k$ along it are given by:
\begin{equation}
  u_k = \frac{X^{\top} v_k}{\lVert X^{\top} v_k \rVert_2},
  \qquad
  s_k = \sqrt{\frac{\lambda_k}{N}}.
  \label{eq:basis}
\end{equation}
Here, the unit vector $u_k$ defines the orientation of structural variation, while the standard deviation $s_k$ provides a principled scaling factor matching the teacher's natural dispersion.\par}

During inference, we perturb the student's intermediate representation using these precomputed directions to induce diversity without modifying model weights. For each generated sample $i$, we independently draw scaling coefficients $\alpha_{i,k} \sim \mathcal{U}[-a, a]$ across the top-$K$ principal directions and inject this calibrated variance directly into the student's activation at the collapsed block, $\tilde{h}_{\ell,1}^{(i)} = h_{\ell,1}^{(i)} + \sum_{k=1}^{K} \alpha_{i,k}\, s_k\, u_k$.

Sampling $\alpha_{i,k}$ independently per generation disperses sampling trajectories into distinct structural modes. We restrict this perturbation strictly to the first denoising step: early steps govern global layout and camera perspective, whereas subsequent steps refine fine-grained textures. Gating the intervention to step 1 steers structural mode selection while allowing subsequent steps to render outputs with native synthesis fidelity. \methodtf{} introduces no parameter updates and can be applied either to the base student or atop \methodft{}; the teacher is needed only once per prompt, to estimate the directions that all samples of that prompt share.
\section{Experiments}
\label{sec:experiments}
\vspace{-5pt}
\subsection{Evaluation Setup}
\label{sec:experiments_settings}

\paragraph{Implementation Details.}
\Needspace*{8\baselineskip}
\begin{wraptable}[8]{r}{0.45\linewidth}
\vspace{-40pt}
  \vspace{-\intextsep}
  \centering
  \caption{\textbf{Teacher--student design choices.}}
  \label{tab:pairs}
  \tabstyledense\setlength{\tabcolsep}{1pt}%
  \begin{tabular}{@{}l cccc@{}}
    \toprule
                        & SDXL      & SD3-M    & Wan~2.1  & H3 \\
    \midrule
    Teacher             & 2.6B UNet & 2B MMDiT & 14B DiT  & 33B DiT \\
    Steps / CFG         & 50 / 8    & 28 / 7   & 50 / 5   & 50 / -- \\
    \addlinespace[1pt]\cdashline{1-5}[1.2pt/1.2pt]\addlinespace[2pt]
    Student (4 steps)   & DMD2      & TDM      & lightx2v & Turbo \\
    \addlinespace[1pt]\cdashline{1-5}[1.2pt/1.2pt]\addlinespace[2pt]
    Resolution          & $1024^2$ & $1024^2$ & $832{\times}480$ & $1344{\times}768$ \\
    Frames              & --        & --       & 81       & 124 \\
    $\ell$ / $\tau$     & 17 / 7    & 12 / 9   & 28 / 8   & 45 / 16 \\
    \addlinespace[1pt]\cdashline{1-5}[1.2pt/1.2pt]\addlinespace[2pt]
    FT: LoRA            & \multicolumn{4}{c}{rank 32, attention + FFN, blocks $\le \ell$} \\
    FT: data            & \multicolumn{4}{c}{100 captions $\times$ $M{=}16$ noises} \\
    FT: training        & \multicolumn{4}{c}{200 iterations, batch 32} \\
    TF: directions      & \multicolumn{4}{c}{top $K{=}8$ from $N{=}25$ teacher runs, $a{=}2$} \\
    \bottomrule
  \end{tabular}
  \vspace{-0.5\intextsep}
\end{wraptable}

Our evaluation spans four teacher--student pairs across two modalities, with U-Net \citep{ronneberger2015u} and DiT \citep{peebles2023scalable} backbones (Tab.~\ref{tab:pairs}, which also gives each pair's collapsed block $\ell$ and teacher step $\tau$); all students are the publicly released 4-step checkpoints, none re-distilled. \emph{\methodft{}} trains LoRA adapters \citep{hu2022lora} on DiffusionDB captions \citep{wang2023diffusiondb} with one recipe for all four pairs and is active only at the first denoising step; \emph{\methodtf{}} estimates the teacher directions of Eq.~(\ref{eq:basis}) per prompt and injects them at the first step. Their settings are in Tab.~\ref{tab:pairs}, details in App.~\ref{app:implementation}.

\vspace{-8pt}
\paragraph{Baselines.}
Besides the teacher and student models, we compare to (i) for images, Noise Optimization \citep{harrington2026s}, which searches for initial noises that maximize a diversity score at test time, and DP-DMD \citep{wu2026diversitypreserved}, which distills a new student whose first step is regressed toward the teacher's latent at an earlier point of its trajectory, and thus needs a full distillation run; and (ii) for videos, DFD \citep{chen2026dataforcing}, which post-trains the student on real videos and thus needs real data and further training (reproduction details in App.~\ref{app:baselines}). Since the teacher is the original multi-step model and Noise Optimization directly maximizes the diversity score, we report both separately as upper bounds, whereas \method{} needs only a short fine-tuning (\methodft{}), and \methodtf{} is training-free, using the teacher's directions without any optimization.

\vspace{-8pt}
\paragraph{Dataset and Evaluation Metrics.}
\Needspace*{5\baselineskip}
\begin{wraptable}{r}{0.34\linewidth}
  \vspace{-\intextsep}
  \centering
  \caption{\textbf{Evaluation data.}}
  \label{tab:eval}
  \tabstyledense
  \begin{tabularx}{\linewidth}{@{}l >{\raggedright\arraybackslash}X >{\raggedright\arraybackslash}X@{}}
    \toprule
                   & Images        & Video \\
    \midrule
    Prompts        & GenEval (553) & VBench (300) \\
    Seeds / prompt & 100           & 10 \\
    \bottomrule
  \end{tabularx}
  \vspace{-0.5\intextsep}
\end{wraptable}

Tab.~\ref{tab:eval} lists the prompts, from GenEval \citep{ghosh2023geneval} and VBench \citep{huang2024vbench}; all arms of a backbone share one noise bank, so every comparison is paired. A \emph{diversity} scorer takes all samples of a prompt and returns the mean pairwise distance between their embeddings, one minus cosine similarity (DINO \citep{caron2021emerging,oquab2024dinov}, DreamSim \citep{fu2023dreamsim}, TPIPS\citep{wang2026tpips}, V-JEPA~2 \citep{assran2025v}, ViCLIP\citep{wang2024internvideo2}) or a perceptual distance (LPIPS \citep{zhang2018unreasonable}); video frames are also compared across clips frame by frame, Vendi \citep{friedman2023the} returns the effective number of distinct samples, and camera-trajectory diversity compares estimated camera paths \citep{chen2026dataforcing}. A \emph{quality or alignment} scorer takes one sample and its prompt and returns a CLIP image--text cosine (CLIPScore \citep{radford2021learning}), a learned preference reward (HPSv2 \citep{wu2023human}, ImageReward \citep{xu2023imagereward}) or a VBench quality dimension. Diversity is averaged over prompts, quality and alignment over samples (App.~\ref{app:protocol}).

\subsection{Results and Analysis}
\label{sec:experiments_results}

\begin{table*}[t]
\caption{\textbf{Main comparison results.} Diversity and quality of distilled image and video generators. \textbf{Bold} and \underline{underline} mark the highest and second-highest value below the dashed line. All metrics are higher-is-better. DS: DreamSim; LP: LPIPS; TP: TPIPS averaged over its aspects; CS: CLIPScore; HPS: HPSv2; IR: ImageReward; VJ: V-JEPA~2; Cam: camera-trajectory diversity; VC: ViCLIP diversity; SC, BC: VBench subject and background consistency; VCS: ViCLIP text--video similarity.}
\label{tab:main_results}
\centering
\tabstyle
\begin{subtable}[c]{0.500\linewidth}
\centering
\caption{Distilled image models.}\vspace{-3pt}
\label{tab:main_image}
\resizebox{\linewidth}{!}{%
\begin{tabular}{@{}l ccccc ccc@{}}
\toprule
 & \multicolumn{5}{c}{\textbf{Diversity}} & \multicolumn{3}{c}{\textbf{Quality}} \\
\cmidrule(lr){2-6} \cmidrule(l){7-9}
\multicolumn{1}{c}{\textbf{Method}} & \textbf{DINO} & \textbf{DS} & \textbf{LP} & \textbf{TP} & \textbf{Vendi} & \textbf{CS} & \textbf{HPS} & \textbf{IR} \\
\midrule
\rowcolor{BlockGray}
\multicolumn{9}{@{}l}{{\scriptsize\bfseries\itshape SDXL}} \\
\tg{Teacher} & \tg{66.3} & \tg{36.0} & \tg{65.9} & \tg{24.1} & \tg{7.66} & \tg{32.9} & \tg{29.2} & \tg{0.695} \\
Noise Opt. & \tg{67.0} & \tg{37.9} & \tg{71.2} & \tg{25.1} & \tg{7.83} & \tg{32.7} & \tg{29.7} & \tg{0.714} \\
\addlinespace[1pt]\cdashline{1-9}[1.2pt/1.2pt]\addlinespace[2pt]
Student & 59.4 & 27.6 & 65.2 & 19.0 & 5.02 & 33.0 & \textbf{30.4} & \textbf{0.854} \\
DP-DMD & 58.3 & 22.7 & 64.1 & 17.6 & 4.32 & 32.6 & 30.2 & 0.709 \\
\addlinespace[2pt]
\rowcolor{OursGreen} \ourstest{} & 60.1 & 28.4 & 65.6 & 19.6 & 5.30 & \textbf{33.0} & \underline{30.3} & \underline{0.836} \\
\rowcolor{OursGreen} \ourstrain{} & \underline{64.6} & \underline{32.9} & \underline{67.8} & \underline{22.6} & \underline{6.74} & 32.8 & 29.8 & 0.694 \\
\rowcolor{OursGreen} \ours{} & \textbf{65.0} & \textbf{33.3} & \textbf{67.9} & \textbf{23.0} & \textbf{7.03} & 32.7 & 29.7 & 0.672 \\
\midrule
\rowcolor{BlockGray}
\multicolumn{9}{@{}l}{{\scriptsize\bfseries\itshape SD3}} \\
\tg{Teacher} & \tg{61.6} & \tg{30.3} & \tg{67.0} & \tg{21.1} & \tg{6.02} & \tg{33.4} & \tg{29.5} & \tg{1.017} \\
Noise Opt. & \tg{56.9} & \tg{26.4} & \tg{61.2} & \tg{18.3} & \tg{6.74} & \tg{33.1} & \tg{29.8} & \tg{0.967} \\
\addlinespace[1pt]\cdashline{1-9}[1.2pt/1.2pt]\addlinespace[2pt]
Student & 48.4 & 19.2 & 55.1 & 13.8 & 3.78 & \underline{33.3} & \underline{30.1} & \textbf{1.026} \\
DP-DMD & \underline{56.8} & 22.4 & \underline{62.2} & \underline{17.4} & 4.46 & 32.5 & \textbf{30.5} & \underline{0.993} \\
\addlinespace[2pt]
\rowcolor{OursGreen} \ourstest{} & 53.8 & \underline{23.5} & 60.7 & 17.0 & \underline{4.86} & 33.1 & 29.7 & 0.965 \\
\rowcolor{OursGreen} \ourstrain{} & 53.5 & 22.7 & 58.0 & 15.3 & 4.66 & \textbf{33.5} & 29.2 & 0.992 \\
\rowcolor{OursGreen} \ours{} & \textbf{60.1} & \textbf{28.3} & \textbf{64.9} & \textbf{20.3} & \textbf{6.24} & 32.9 & 29.0 & 0.872 \\
\bottomrule
\end{tabular}}
\end{subtable}%
\hfill
\begin{subtable}[c]{0.484\linewidth}
\centering
\caption{Distilled video models.}\vspace{-3pt}
\label{tab:main_video}
\resizebox{\linewidth}{!}{%
\begin{tabular}{@{}l ccccc ccc@{}}
\toprule
 & \multicolumn{5}{c}{\textbf{Diversity}} & \multicolumn{3}{c}{\textbf{Quality}} \\
\cmidrule(lr){2-6} \cmidrule(l){7-9}
\multicolumn{1}{c}{\textbf{Method}} & \textbf{DINO} & \textbf{VJ} & \textbf{TP} & \textbf{Cam} & \textbf{VC} & \textbf{SC} & \textbf{BC} & \textbf{VCS} \\
\midrule
\rowcolor{BlockGray}
\multicolumn{9}{@{}l}{{\scriptsize\bfseries\itshape Wan 2.1}} \\
\addlinespace[2pt]
\tg{Teacher} & \tg{51.6} & \tg{20.9} & \tg{31.3} & \tg{36.2} & \tg{36.2} & \tg{94.7} & \tg{97.3} & \tg{21.4} \\
\addlinespace[1pt]\cdashline{1-9}[1.2pt/1.2pt]\addlinespace[2pt]
Student & 31.6 & 11.2 & 23.2 & 36.2 & 22.6 & \underline{96.7} & 95.6 & 21.5 \\
\addlinespace[2pt]
DFD & 26.2 & 10.0 & 17.8 & 25.1 & 18.5 & \textbf{97.4} & 96.4 & \textbf{22.0} \\
\addlinespace[2pt]
\rowcolor{OursGreen} \ourstest{} & 34.6 & 12.5 & 26.3 & 35.8 & 26.1 & 96.3 & 95.2 & \underline{21.5} \\
\rowcolor{OursGreen} \ourstrain{} & \underline{46.8} & \underline{17.0} & \underline{28.7} & \textbf{40.6} & \underline{34.7} & 96.4 & \textbf{96.4} & 20.8 \\
\rowcolor{OursGreen} \ours{} & \textbf{52.2} & \textbf{18.7} & \textbf{32.6} & \underline{40.0} & \textbf{40.0} & 95.0 & 95.9 & 20.5 \\
\midrule
\rowcolor{BlockGray}
\multicolumn{9}{@{}l}{{\scriptsize\bfseries\itshape MiniMax-H3}} \\
\addlinespace[2pt]
\tg{Teacher} & \tg{52.5} & \tg{20.7} & \tg{30.4} & \tg{37.7} & \tg{37.2} & \tg{92.8} & \tg{94.6} & \tg{20.7} \\
\addlinespace[1pt]\cdashline{1-9}[1.2pt/1.2pt]\addlinespace[2pt]
Student & 36.5 & 12.4 & 23.0 & 26.9 & 28.4 & 94.0 & 95.1 & 19.9 \\
\addlinespace[2pt]
DFD & 35.6 & 12.3 & 22.2 & 26.7 & 27.2 & \underline{94.3} & \textbf{95.8} & 20.1 \\
\addlinespace[2pt]
\rowcolor{OursGreen} \ourstest{} & 42.2 & 13.4 & \underline{26.4} & 29.4 & \underline{33.6} & 93.4 & 94.1 & 19.8 \\
\rowcolor{OursGreen} \ourstrain{} & \underline{44.1} & \underline{15.9} & 26.2 & \underline{33.2} & 32.2 & \textbf{94.8} & \underline{95.5} & \textbf{20.1} \\
\rowcolor{OursGreen} \ours{} & \textbf{50.9} & \textbf{16.2} & \textbf{29.9} & \textbf{35.6} & \textbf{38.6} & 94.2 & 94.0 & 19.8 \\
\bottomrule
\end{tabular}}
\end{subtable}
\end{table*}
 
\begin{figure}[t]
  \centering
  \includegraphics[width=\linewidth]{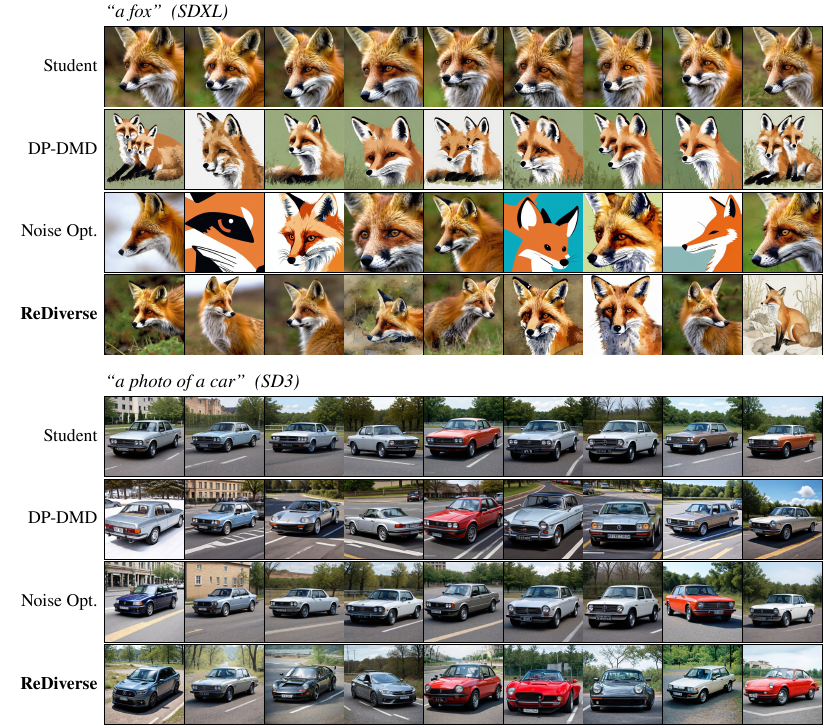}
  \caption{\textbf{Qualitative results on SDXL and SD3.} One prompt per backbone and four 4-step
    methods are shown across nine columns. Each column corresponds to one
    initial noise, shared across all four methods.}
  \label{fig:qualitative_image}
\end{figure}

\begin{wrapfigure}{r}{0.46\linewidth}
  \vspace{-10pt}
  \centering
  \includegraphics[width=\linewidth]{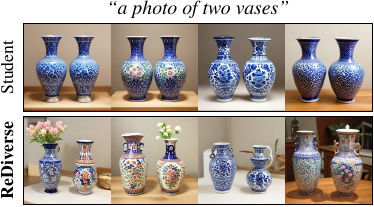}
  \vspace{-6pt}
  \caption{\textbf{Diversity within an image.} The SD3 student renders the same
    vase twice, whereas \method{} increases the diversity while keeping the same count. Same column, same noise.}
  \label{fig:intra}
  \vspace{-8pt}
\end{wrapfigure}

\paragraph{\method{} improves diversity on images.} Tab.~\ref{tab:main_image} compares the 4-step methods on SDXL and SD3. Both students lose substantial diversity relative to their teachers, while \method{} recovers most of this gap and achieves the highest diversity among 4-step methods on every metric for both backbones. \methodft{} and \methodtf{} each recover part of the gap, with their combination providing the largest gain. \method{} even exceeds the teacher on LPIPS for SDXL and Vendi for SD3, and surpasses Noise Optimization on all diversity metrics except Vendi for SD3. DP-DMD falls below the student on SDXL and provides smaller gains than \method{} on SD3. CLIPScore changes by less than 0.5 points from the student, while the reward scores remain close to the teacher on SDXL. Qualitative results are shown in Fig.~\ref{fig:teaser}, Fig.~\ref{fig:qualitative_image} and App.~\ref{app:qual_images}, and Fig.~\ref{fig:intra} shows the diversity within individual images.

\begin{figure}[t]
  \centering
  \includegraphics[width=\linewidth]{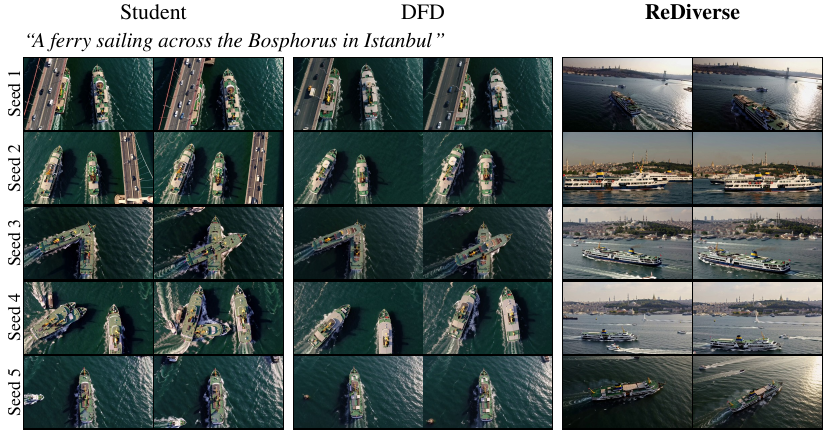}
  \caption{\textbf{Qualitative results on MiniMax-H3.} Rows are seeds, each clip shown by its first and last
    frame, same noise for all three columns. Best viewed as videos available in supplementary material.}
  \label{fig:qualitative}
\end{figure}

\begin{figure}[t]
  \centering
  \includegraphics[width=\linewidth]{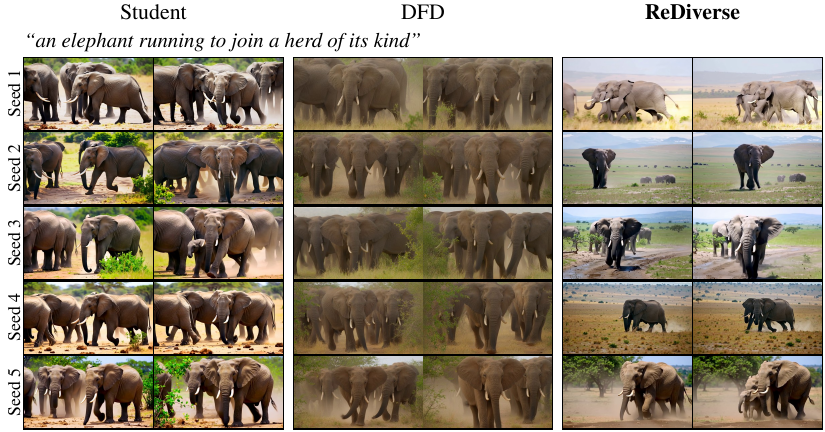}
  \caption{\textbf{Qualitative results on Wan~2.1.} Rows are seeds, each clip shown by its first and last
    frame, same noise for all three columns. Best viewed as videos available in supplementary material.}
  \label{fig:qualitative_wan_elephant}
\end{figure}

\vspace{-8pt}
\paragraph{\method{} improves diversity on video.}
Tab.~\ref{tab:main_video} compares the 4-step methods on Wan~2.1 and MiniMax-H3. Both students show a large drop in diversity relative to their teachers, and \method{} recovers most of this gap. On Wan~2.1, \method{} matches or exceeds the teacher on all diversity metrics except V-JEPA~2, while on MiniMax-H3 it approaches the teacher and surpasses it on ViCLIP diversity. The two students also exhibit different diversity failures: the Wan~2.1 student largely preserves camera-trajectory diversity but loses content diversity, whereas the H3 student loses both. \method{} recovers these missing variations in both cases, without explicitly targeting any particular type of variation, by matching the teacher's activation structure. In contrast, DFD reduces diversity on Wan~2.1 and provides little change on MiniMax-H3. This diversity recovery does not substantially affect quality: subject and background consistency remain close to the teacher, while text--video similarity stays within one point of the student. Seed-matched examples are shown in Fig.~\ref{fig:teaser}, Fig.~\ref{fig:qualitative} (MiniMax-H3) and Fig.~\ref{fig:qualitative_wan_elephant} (Wan~2.1), with additional results for both video backbones in App.~\ref{app:qual_video}.

\raggedbottom   %
\Needspace*{22\baselineskip}   %

\vspace{-8pt}
\paragraph{Diversity within an image.}
Diversity collapse also occurs within individual images: when a prompt requires multiple instances of the same object, the student often represents them with nearly the same appearance. The directional collapse explains this behavior: with fewer available activation directions, different spatial positions have less freedom to represent distinct instances and tend to fall into the same mode. We tested this on ``a photo of two vases'' and found that ReDiverse restores distinct appearances for the two vases while preserving their count, layout, and framing (Fig.~\ref{fig:intra}). Thus, restoring the missing directions improves diversity not only across samples, but also between instances within a single image. 
\flushbottom

\vspace{-8pt}
\paragraph{User and VLM-as-a-Judge Study.}
\begin{figure}[!h]
  \centering
  \includegraphics[width=\linewidth]{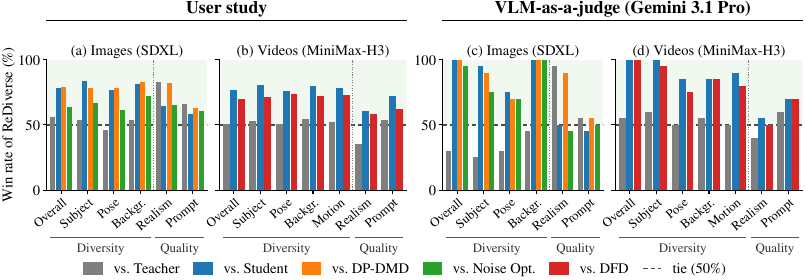}
  \caption{\textbf{User and VLM-as-a-judge study.} Win rate of \method{} against each method on diversity (overall and per aspect: subject,
  pose, background, and motion for videos), realism and prompt following; a tie counts half, and
  above the dashed 50\% line \method{} is preferred.}
  \label{fig:human_study}
\end{figure}

Automatic metrics provide an indirect measure of diversity, so we also evaluate whether people actually perceive the samples as more diverse. This is particularly important for videos, where diversity also involves differences in motion. We therefore ask raters to compare \method{} with one method at a time, using the same prompt and initial noises, and judge which is more diverse (overall and by aspect), more realistic, and more faithful to the prompt. Gemini 3.1 Pro \citep{google2026gemini31pro} performs the same comparisons as an automatic judge (App.~\ref{app:user_study}). In the user study (Fig.~\ref{fig:human_study}), raters prefer \method{} over every 4-step method in diversity, both overall and across all aspects, for both images and videos. They rate \method{} on par with the teacher, even slightly higher overall, and also prefer it in prompt following against every method. The VLM judge agrees that \method{} is more diverse than the 4-step methods (Fig.~\ref{fig:human_study}). More details about the user study are provided in App.~\ref{app:user_study}.

\vspace{-8pt}
\paragraph{Ablation Study.}
\newcommand{\bandstrut}{\rule[-0.8ex]{0pt}{2.9ex}}
\definecolor{DeltaUp}{RGB}{34,139,34}\definecolor{DeltaDown}{RGB}{200,40,40}
\ifdefined\dltw\else\newlength{\dltw}\fi\settowidth{\dltw}{\scriptsize +00.0}
\providecommand{\dbox}[1]{\,\makebox[\dltw][l]{\scriptsize #1}}
\providecommand{\dup}[1]{\dbox{\textcolor{DeltaUp}{+#1}}}
\providecommand{\ddn}[1]{\dbox{\textcolor{DeltaDown}{$-$#1}}}
\providecommand{\dzr}{\dbox{\textcolor{TextGray}{$\pm$0.0}}}
\providecommand{\dnone}{\dbox{}}
\ifdefined\dltwr\else\newlength{\dltwr}\fi\settowidth{\dltwr}{\scriptsize $-$0.000}
\providecommand{\dboxr}[1]{\,\makebox[\dltwr][l]{\scriptsize #1}}
\providecommand{\dupr}[1]{\dboxr{\textcolor{DeltaUp}{+#1}}}
\providecommand{\ddnr}[1]{\dboxr{\textcolor{DeltaDown}{$-$#1}}}
\providecommand{\dnoner}{\dboxr{}}
\begin{table}[t]
\caption{\textbf{Ablation on SD3.} Sand rows: defaults; small numbers: change from the default.}
\label{tab:ablation}
\centering
\tabstyle
\begin{subtable}[c]{0.498\linewidth}
\centering
\caption{\methodft{}: where and when to regress.}\vspace{-3pt}
\resizebox{\linewidth}{!}{%
\begin{tabular}{@{}l ccc@{}}
\toprule
\textbf{Setting} & \textbf{DINO}$\uparrow$ & \textbf{TPIPS}$_{\text{avg}}\uparrow$ & \textbf{IR}$\uparrow$ \\
\midrule
\tg{Teacher} & \tg{61.0}\dnone & \tg{21.7}\dnone & \tg{0.885}\dnoner \\
\tg{Student} & \tg{46.9}\dnone & \tg{13.9}\dnone & \tg{0.955}\dnoner \\
\midrule
Earlier block: $8$ & 48.9\ddn{2.7} & 14.6\ddn{0.8} & 0.921\ddnr{0.013} \\
Later block: $16$ & 55.4\dup{3.8} & 17.0\dup{1.6} & 0.839\ddnr{0.095} \\
Regress at all blocks & 57.8\dup{6.2} & 26.6 & 0.801\ddnr{0.133} \\
\addlinespace[1pt]\cdashline{1-4}[1.2pt/1.2pt]\addlinespace[2pt]
Regressed steps: $t{=}1$--$4$ & 50.0\ddn{1.6} & 15.6\dup{0.2} & $-$0.564\ddnr{1.498} \\
Adapter steps: $t{=}1$--$4$ & 51.7\dup{0.1} & 16.1\dup{0.7} & 0.380\ddnr{0.554} \\
\addlinespace[1pt]\cdashline{1-4}[1.2pt/1.2pt]\addlinespace[2pt]
\rowcolor{AblPick} \methodft{} ($\ell{=}12$, $t{=}1$) & 51.6\dnone & 15.4\dnone & 0.934\dnoner \\
\bottomrule
\end{tabular}}
\end{subtable}%
\hfill
\begin{subtable}[c]{0.486\linewidth}
\centering
\caption{\methodtf{}: which directions, and how many.}\vspace{-3pt}
\resizebox{\linewidth}{!}{%
\begin{tabular}{@{}l ccc@{}}
\toprule
\textbf{Setting} & \textbf{DINO}$\uparrow$ & \textbf{TPIPS}$_{\text{avg}}\uparrow$ & \textbf{IR}$\uparrow$ \\
\midrule
\tg{Teacher} & \tg{61.0}\dnone & \tg{21.7}\dnone & \tg{0.885}\dnoner \\
\tg{Student} & \tg{46.9}\dnone & \tg{13.9}\dnone & \tg{0.955}\dnoner \\
\midrule
Student's own top $K{=}8$ & 49.5\ddn{3.5} & 14.7\ddn{2.6} & 0.954\dupr{0.076} \\
Random $K{=}8$, teacher's $s_k$ & 57.8\dup{4.8} & 18.4\dup{1.1} & 0.719\ddnr{0.159} \\
\addlinespace[1pt]\cdashline{1-4}[1.2pt/1.2pt]\addlinespace[2pt]
Teacher's top $K{=}1$ & 48.2\ddn{4.8} & 14.6\ddn{2.7} & 0.941\dupr{0.063} \\
\addlinespace[1pt]\cdashline{1-4}[1.2pt/1.2pt]\addlinespace[2pt]
\rowcolor{AblPick} \methodtf{} ($K{=}8$, $a{=}2$) & 53.0\dnone & 17.3\dnone & 0.878\dnoner \\
\bottomrule
\end{tabular}}
\end{subtable}
\end{table}
Tab.~\ref{tab:ablation} changes one design choice at a time on SD3 (full sweeps in App.~\ref{app:design}). (i)~\methodft{} should regress where the teacher's directions are lost, at the collapsed block: at an earlier block the student still matches the teacher, so there is less to correct and the gain shrinks, while at a later block the adapter also reshapes the blocks that render the image, adding diversity at a cost in quality (App.~\ref{app:layer}). As the layout is fixed at the first step, regressing at every step or keeping the adapter on after it adds no diversity and lowers prompt alignment (App.~\ref{app:gate}). (ii)~For \methodtf{}, what matters is which directions receive variance: the student's own directions change appearance but not layout (Fig.~\ref{fig:steering}), a single teacher direction is too few because the lost variance is spread over many axes, and random directions of the same scale make samples differ mostly in background texture, at a cost in prompt alignment (App.~\ref{app:randdir}).
\section{Discussion}
\label{sec:discussion}
Probing the activations of distilled generators shows that diversity is lost at a single collapsed block, where the student's features lose the teacher's principal directions of variation that the downstream blocks can still render. \method{} restores them with a short, localized fine-tuning (\methodft{}) or without training (\methodtf{}), on images and videos. A limitation is that quality still drops at times, most visibly in reward-model scores on images (Tab.~\ref{tab:main_results}) and in video realism against the teacher (Fig.~\ref{fig:human_study}), as \method{} corrects the student after distillation rather than during it. Both variants also assume that the student shares the teacher's architecture, as in step distillation; compressed students such as BK-SDM \citep{kim2024bksdm} would need a learned projection between matched blocks, as in FitNets \citep{romero2015fitnets}. In future work, we will use this probing to design a new distillation method that keeps the teacher's directions at the collapsed block during distillation, so that the student is diverse at the first shot.

\subsection*{Acknowledgments}
We thank fal for the compute grants that supported this research. We acknowledge the support of the Health Care Engineering Systems Center within the Coordinated Science Lab at UIUC.

\section*{Contribution Statement}
\textbf{Furkan Horoz} led and implemented the probing analysis and both ReDiverse variants, and ran all experiments on the four teacher--student pairs, including the baselines, ablations, user study and VLM-as-a-judge evaluation.
\newline
\textbf{Ozgur Kara} guided Furkan Horoz and was closely involved through every step of the project, contributed to shaping the analysis and the experimental design, and prepared the manuscript, including its framing and presentation of the results.
\newline
\textbf{Kaan Oktay} provided technical guidance throughout the project, including feedback on few-step distillation, large-scale image and video generation, and the experimental setup.
\newline
\textbf{Adil Kaan Akan} provided technical guidance throughout the project, contributed to the design and execution of the user study, and gave feedback on the experimental setup.
\newline
\textbf{Atakan Kara} provided technical guidance throughout the project, contributed to the design of the main figure, and gave feedback on few-step distillation and the experimental setup.
\newline
\textbf{Anand Bhattad} shaped the technical direction of the project. He encouraged the team to localize where diversity collapses inside the network, steered the work toward video models, and pushed for restoring diversity along the teacher's principal directions. He also pushed for a more granular evaluation, including TPIPS and comparisons against recent distillation methods for image and video models. He met weekly with FH and OK to give feedback on the analysis, figures and results, wrote the introduction, and revised substantial parts of the manuscript.
\newline
\textbf{James Matthew Rehg} supervised the project as principal investigator of the lab, provided guidance on the research direction and the manuscript, and co-advised the project.

\subsection*{AI Use Statement}
We used generative AI tools to polish the text of the manuscript and to write the code that produces
plots from our own measurements. Additionally, we used generative AI tools for literature search
and retrieval; all retrieved works were read and analysed by the authors. Every AI-assisted passage
was checked, and all AI-assisted code was tested, by the authors. We take responsibility for the final
content of this work, including text, claims or artifacts produced with the aid of generative AI.

\subsection*{Ethics Statement}
This work modifies publicly released text-to-image and text-to-video models. Restoring diversity lets a distilled
model cover the range of outputs its teacher already produces, countering the homogenized outputs of distillation;
it adds no new generative capability, so its potential for misuse and its biases are those of the underlying models.
The user study was completely anonymous: raters only compared generated images and videos, and no personal data
was collected.

\clearpage   %
\bibliography{refs}
\bibliographystyle{iclr2027_conference}

\newif\ifsupp
\supptrue
\ifsupp
\clearpage
\appendix
\makeatletter
\def\addcontentsline#1#2#3{%
  \addtocontents{#1}{\protect\contentsline{#2}{#3}{\thepage}{}\protected@file@percent}}
\makeatother
\etocdepthtag.toc{appendix}
\raggedbottom
\let\origappsection\section
\renewcommand{\section}{\clearpage\origappsection}
\etocsettagdepth{main}{none}
\etocsettagdepth{appendix}{subsection}
\etocsettocstyle{\section*{Supplementary Material}\noindent The supplementary material is organised as follows.\par\medskip}{}
\tableofcontents

\section{Measuring the Variance Characteristic}
\label{app:measurement}

Our starting question was the following: the teacher generates different-looking outputs for different noises, but
the student does not; for the same set of noises it generates similar-looking outputs. Our idea was therefore to look
at the internal representations behind these outputs, to understand how different noises end up as different-looking
outputs.

To measure this, we feed many initial noises through the network, noises that generate different outputs in the
teacher and similar outputs in the student, and look at what they produce inside the network. At an internal block each noise
gives one activation. We flatten each activation into a vector, so from $N$ noises we have
$N$ activation vectors, that is, $N$ points in activation space, and we measure the characteristics of the
distribution formed by these points. We do this at every block of the network. With $X_b \in \mathbb{R}^{N \times D}$
holding the $N$ vectors at block $b$, centred over the noises, the covariance is
\begin{equation}
  \Sigma_b = \frac{1}{N} X_b^{\top} X_b ,
\end{equation}
and its eigenvalues $\lambda_j$ give the variance along each principal direction of the distribution.

\paragraph{How we measure.}
We give the same $N = 1000$ noises to the teacher and to the student as the input of their first denoising step, run
the conditional branch only, and keep the image or video positions of every block. We use 100 prompts for SDXL and
SD3, 20 for Wan~2.1 and MiniMax-H3. Since $D$ is very large, instead of estimating the $D \times D$ covariance
matrix we estimate the $N \times N$ Gram matrix $G_b = X_b X_b^{\top} / N$, whose entries are the inner products
between the centred activations of every pair of noises:
\begin{equation}
  (G_b)_{ij} = \frac{1}{N} \left(h_i - \bar{h}\right)^{\top} \left(h_j - \bar{h}\right) ,
  \qquad
  \bar{h} = \frac{1}{N} \sum_{i=1}^{N} h_i ,
\end{equation}
where $h_i \in \mathbb{R}^{D}$ is the flattened activation of noise $i$ at block $b$. The Gram matrix has the same
nonzero eigenvalues as $\Sigma_b$.

\paragraph{Total variance.}
We first measure the total variance of this distribution, the sum of the eigenvalues:
\begin{equation}
  \mathrm{Tr}(\Sigma_b) = \sum_j \lambda_j .
\end{equation}
It measures how far the points spread, that is, how much the activation moves when the noise changes. The student's
variance is not lost (Fig.~\ref{fig:supp_depth_var}); in fact it is higher than the teacher's at most blocks. This is surprising:
since the student's outputs collapse to a few modes, we expected its activations to collapse in variance as well.

\begin{figure}[h]
  \centering
  \includegraphics[width=\linewidth]{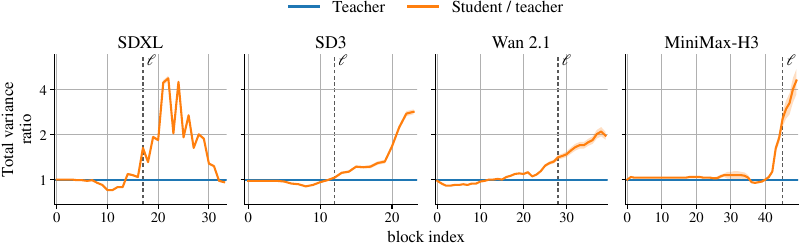}
  \caption{\textbf{Total variance}, student divided by teacher, at every block of the first denoising step; the
  teacher is the line at 1. Mean of the per-prompt ratios with 95\% confidence interval; dashed line: the collapsed
  block $\ell$.}
  \label{fig:supp_depth_var}
\end{figure}

\paragraph{Participation ratio.}
Since we could not find a variance collapse in the internal blocks of the network, we started to check whether this
variance is collapsed into a few directions or covers many directions. An effective way of measuring this is the
participation ratio, which counts how many directions the spread effectively occupies:
\begin{equation}
  \mathrm{PR}(\Sigma_b) = \frac{\left(\sum_j \lambda_j\right)^{2}}{\sum_j \lambda_j^{2}} .
\end{equation}
A distribution that extends evenly in many directions has a high value, and one stretched along a few directions a
low one. The student follows the teacher in the early blocks and then drops below it in the
intermediate blocks (Fig.~\ref{fig:supp_depth_pr}). So the variance is not lost, but it is
collapsed into fewer directions than in the teacher.

\begin{figure}[h]
  \centering
  \includegraphics[width=\linewidth]{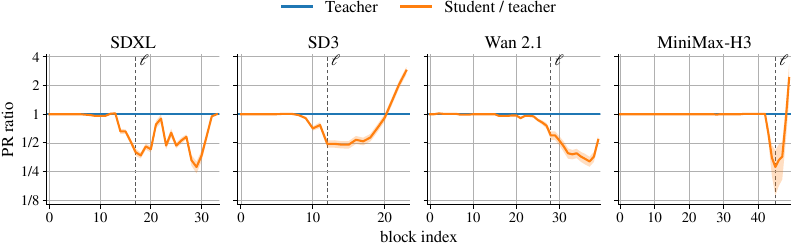}
  \caption{\textbf{Participation ratio}, student divided by teacher, at every block of the first denoising step; the
  teacher is the line at 1. Mean of the per-prompt ratios with 95\% confidence interval; dashed line: the collapsed
  block $\ell$.}
  \label{fig:supp_depth_pr}
\end{figure}

\paragraph{Concentration.}
The student keeps the variance but spreads it over fewer directions, so we asked how much variance each of these
directions carries. We measure this with the concentration, the total variance divided by the participation ratio,
\begin{equation}
  \mathrm{Conc}(\Sigma_b) = \frac{\mathrm{Tr}(\Sigma_b)}{\mathrm{PR}(\Sigma_b)} ,
\end{equation}
and indeed each remaining direction of the student carries more variance than the teacher's
(Fig.~\ref{fig:supp_depth_conc}).

\begin{figure}[h]
  \centering
  \includegraphics[width=\linewidth]{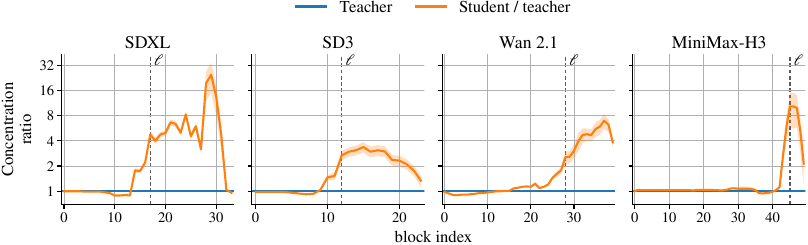}
  \caption{\textbf{Concentration}, student divided by teacher, at every block of the first denoising step; the
  teacher is the line at 1. Mean of the per-prompt ratios with 95\% confidence interval; dashed line: the collapsed
  block $\ell$.}
  \label{fig:supp_depth_conc}
\end{figure}

\paragraph{Choosing the collapsed block.}
The drop in participation ratio does not happen at a single block; it builds up over a few blocks
(Fig.~\ref{fig:supp_depth_pr}). We choose the collapsed block inside this region, at the block where the
concentration makes a clear step up (Fig.~\ref{fig:supp_depth_conc}): there the student's variance has moved into
fewer directions, each carrying much more variance than in the teacher, and in almost all following blocks the
student stays more concentrated than the teacher. This is also the block that \methodft{} regresses, and our
ablation on SD3 supports this choice: regressing earlier constrains a representation that has not lost its
directions yet and recovers less diversity, while regressing later recovers more diversity but also changes the
blocks that render the image and lowers the output quality (App.~\ref{app:layer}).

\paragraph{Eigenvalues at the collapsed block.}
We can also see this directly in the eigenvalues. Fig.~\ref{fig:supp_eigshare} shows, at the collapsed block, which
percentage of the total variance each of the leading eigenvalues carries. In the student, the leading eigenvalues carry
a much larger share of the variance than in the teacher on all four models, and the first ten of them carry a larger
share than the teacher's on every prompt: most of the student's variance sits in a few directions, while the teacher
spreads it more evenly over many.

\begin{figure}[h]
  \centering
  \includegraphics[width=\linewidth]{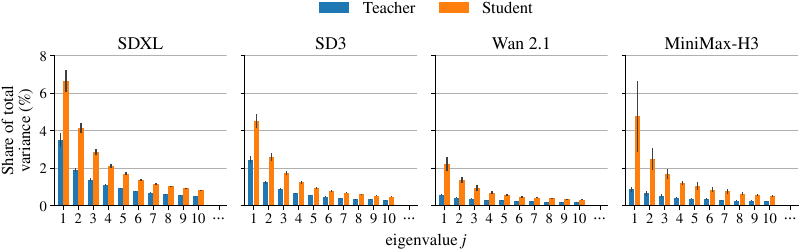}
  \caption{\textbf{Share of the total variance per eigenvalue} at the collapsed block, first denoising step, for the 10
  largest eigenvalues. Mean over prompts; whiskers: 95\% confidence interval.}
  \label{fig:supp_eigshare}
\end{figure}

\paragraph{The next steps.}
So far we compared the teacher and the student at the first denoising step. Since one student step replaces several
teacher steps, we also ran the teacher along its own denoising steps from the same noises and measured its
participation ratio and concentration at the collapsed block at each of its steps up to step $\tau$, the teacher step
whose activation \methodft{} regresses ($\tau = 7$ for SDXL and 8 for Wan~2.1), against the student's first step
(Fig.~\ref{fig:supp_steps}). Along its steps the teacher's
participation ratio decreases and its concentration increases, but the student's first step still has, on average,
fewer directions and a higher concentration than the teacher at every step up to $\tau$.

\begin{figure}[h]
  \centering
  \includegraphics[width=\linewidth]{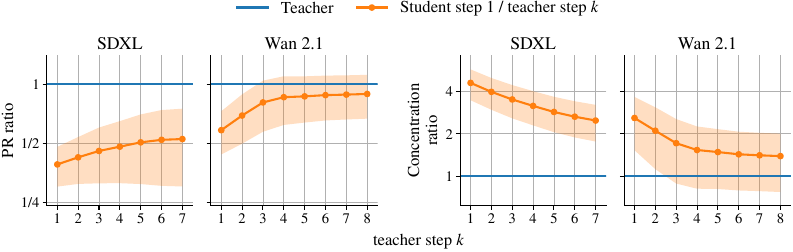}
  \caption{\textbf{Participation ratio and concentration along the teacher's steps} at the collapsed block: the
  student's first step divided by the teacher's step $t$, for the teacher's first $\tau$ steps; the teacher is the line
  at 1. Mean of the per-prompt
  ratios with 95\% confidence interval.}
  \label{fig:supp_steps}
\end{figure}

\paragraph{After the fine-tuning.}
\methodft{} regresses the student's first-step activation at the collapsed block onto the teacher's, so the
quantity it should move is the one measured above. We repeat the measurement on SDXL and Wan~2.1 for the student
carrying the adapter, at the first denoising step and against the teacher's first step. The span at the
collapsed block widens: the participation ratio rises from $120$ to $190$ on SDXL and from $535$ to $593$ on
Wan~2.1, against $317$ and $918$ for the teacher.
\section{Steering: Protocol and Further Examples}
\label{app:steering}

In Sec.~\ref{sec:diagnosis_steering} we asked whether the directions the student loses are the ones that make the
outputs different, and whether the student can still use them. We take one noise, move its activation at the collapsed
block along a single direction, and look at how the output changes: if its layout changes, this direction carries the
diversity the student is missing; if only small details change, it does not.

\paragraph{Directions.}
For a prompt we feed $N{=}25$ noises, other than the one we steer, to a network and stack their activations at the
collapsed block, centred over the noises, as the rows of $X \in \mathbb{R}^{N \times D}$. Since $D$ is very large, we use
the Gram matrix $X X^{\top}$ instead of the covariance matrix: with its top eigenvalue $\lambda_1$ and eigenvector $v_1$,
the top principal direction and the spread of the noises along it are
\begin{equation}
  u_1 = \frac{X^{\top} v_1}{\lVert X^{\top} v_1 \rVert_2}, \qquad s_1 = \sqrt{\lambda_1 / N},
\end{equation}
where $s_1$ is the standard deviation of the noises along $u_1$. We compute these directions for both the
teacher and the student, to check how each model responds to its own direction and how the student responds to the
teacher's direction.

\paragraph{Steering.}
We take one noise from our evaluation noise bank and, during the first forward pass only, replace the activation $h$ at
the collapsed block by $h + \alpha\, s_1 u_1$, leaving the rest of the sampling unchanged; the dose $\alpha$ is in units
of $s_1$. The teacher is steered along its own direction with its usual guidance, the shift added to its conditional
branch; the student is steered along the teacher's direction and along its own. We find that steering the student along
the teacher's direction creates diversity, while steering it along its own direction does not (Fig.~\ref{fig:steering_supp}):
along the teacher's direction the cat close-up opens into a view of the cat in a room, along its own only colour,
contrast and lighting change.

\begin{figure}[h]
  \centering
  \includegraphics[width=\linewidth]{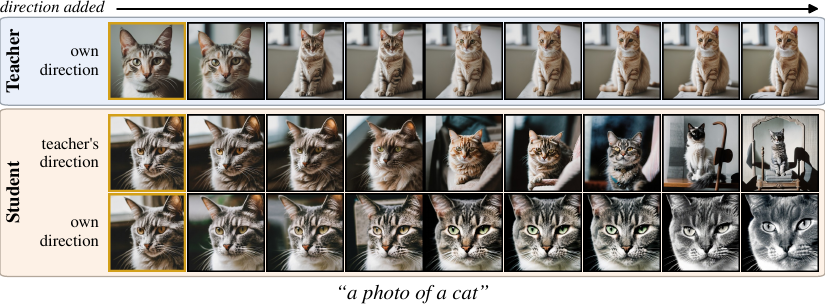}
  \caption{\textbf{Steering on SDXL.} The top box moves the teacher along its own direction and the
  bottom box moves the student along the teacher's direction and along its own direction. Same noise in every row;
  gold frame: no steering; the dose grows from left to right.}
  \label{fig:steering_supp}
\end{figure}
\section{Two Students of One Teacher}
\label{app:same_teacher}

In Sec.~\ref{sec:diagnosis_steering} we saw that two students of the same teacher, MiniMax-H3, reach different
output diversity, and that the more diverse one keeps more directions at the collapsed block.

\paragraph{Students.}
Both are 4-step students of MiniMax-H3, used as released. FastVideo \citep{fastvideo2026} distills all weights of the
teacher with data-free DMD2; Turbo \citep{lightx2v2026turbo} is a LoRA on the teacher's weights.

\paragraph{Measurement.}
As in App.~\ref{app:measurement}, for each prompt we feed the same 1000 noises to the teacher and to both students and
compute the total variance, PR and concentration at every block of the first step, on the conditional branch only; we
use 5 prompts and divide each student's value by the teacher's. For diversity, every model generates 5 clips for each
of 20 VBench prompts from the same noises.

\begin{figure}[h]
  \centering
  \includegraphics[width=\linewidth]{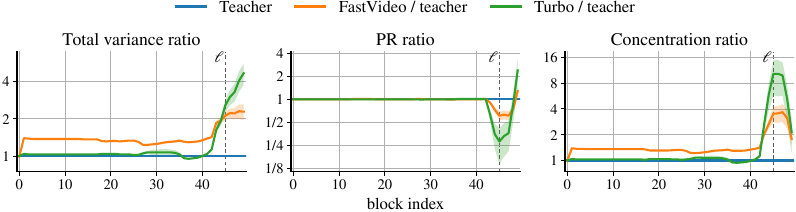}
  \caption{\textbf{Two students of MiniMax-H3 along depth.} Total variance, PR and concentration of each student
  divided by the teacher's at every block of the first step; mean over 5 prompts with 95\% confidence interval; dashed
  line: the collapsed block $\ell$ (block 45).}
  \label{fig:supp_same_teacher}
\end{figure}

\paragraph{Results.}
Both students keep the teacher's PR until block 42 (Fig.~\ref{fig:supp_same_teacher}). From block 43 on both lose
directions, but Turbo collapses much more than FastVideo: its PR drops further and its variance is concentrated on
fewer directions. The outputs follow the same order (Tab.~\ref{tab:supp_same_teacher}, Fig.~\ref{fig:supp_same_teacher_qual}):
FastVideo is more diverse than Turbo on every diversity metric and close to the teacher.

\begin{table}[h]
\centering
\caption{\textbf{Two students of MiniMax-H3.} PR at the collapsed block divided by the teacher's, measured on
the 5 prompts of this section; Tab.~\ref{tab:pr_summary} reports $0.37$ for the same student on the 20 prompts of
App.~\ref{app:measurement}. Diversity and
text alignment on 20 VBench prompts with 5 clips each. All scores $\times 100$ except Vendi.}
\label{tab:supp_same_teacher}
\small
\setlength{\tabcolsep}{5pt}
\begin{tabular}{lc ccccc c}
\toprule
& & \multicolumn{5}{c}{\textbf{Diversity}} & \textbf{Alignment} \\
\cmidrule(lr){3-7} \cmidrule(l){8-8}
\textbf{Model} & \textbf{PR ratio at $\ell$} & \textbf{DINO}$\uparrow$ & \textbf{V-JEPA2}$\uparrow$ &
\textbf{TPIPS}$_{\text{avg}}\uparrow$ & \textbf{ViCLIP}$_{\text{div}}\uparrow$ & \textbf{Vendi}$\uparrow$ &
\textbf{ViCLIP-S}$\uparrow$ \\
\midrule
\textcolor{gray}{Teacher} & \textcolor{gray}{1.00} & \textcolor{gray}{55.3} & \textcolor{gray}{21.8} &
\textcolor{gray}{31.0} & \textcolor{gray}{39.9} & \textcolor{gray}{2.77} & \textcolor{gray}{19.7} \\
FastVideo & 0.61 & \textbf{52.3} & \textbf{17.5} & \textbf{30.1} & \textbf{39.0} & \textbf{2.74} & 18.8 \\
Turbo     & 0.28 & 35.8 & 11.3 & 22.9 & 27.7 & 2.21 & \textbf{20.3} \\
\bottomrule
\end{tabular}
\end{table}

\begin{figure}[h]
  \centering
  \includegraphics[width=\linewidth]{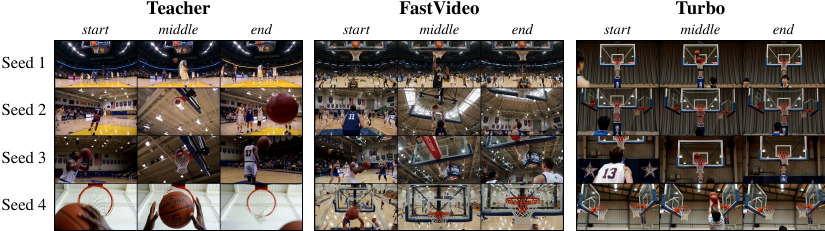}
  \caption{\textbf{``A person is dunking basketball'' on MiniMax-H3.} Start, middle and end frame of 4 clips per
  model, same noise in each row.}
  \label{fig:supp_same_teacher_qual}
\end{figure}
\section{Implementation Details}
\label{app:implementation}

This section provides the implementation details underlying
Sec.~\ref{sec:experiments_settings}. We describe the teacher and student
checkpoints, the collapsed block used for each model pair, the training
procedure for \methodft{}, and the estimation and injection of directions
used by \methodtf{}.

\subsection{Backbones and Collapsed Blocks}
\label{app:impl_backbones}

We use the publicly released 4-step student checkpoints for all four
backbones without modification (Tab.~\ref{tab:impl_backbones}). The collapsed
block $\ell$ is selected as described in App.~\ref{app:measurement}, and
$\tau$ denotes the teacher step whose activation is used as the regression
target for \methodft{}.

\begin{table}[h]
\centering
\caption{\textbf{Checkpoints and collapsed blocks.} Teacher and student
checkpoints, the teacher sampler, the collapsed block $\ell$, the teacher
step $\tau$ used for the regression target, and the guidance used when
computing the target.}
\label{tab:impl_backbones}
\scriptsize
\setlength{\tabcolsep}{3pt}
\begin{tabular}{@{}l llll@{}}
\toprule
& SDXL & SD3-M & Wan~2.1 & MiniMax-H3 \\
\midrule
Teacher & SDXL-base-1.0 & SD3-medium & Wan2.1-T2V-14B & MiniMax-H3 \\
Student & DMD2 (4-step UNet) & TDM (LoRA) & lightx2v (step- and CFG-distilled) & lightx2v Turbo \\
Teacher sampler & Euler, 50 steps & FM-Euler, 28 steps & UniPC, 50 steps & FM, 50 steps \\
Collapsed block $\ell$ & \texttt{up0.attn0} & \texttt{block 12} & \texttt{block 28} & \texttt{block 45} \\
Teacher step $\tau$ & 7 & 9 & 8 & 16 \\
Teacher guidance for the target & CFG 8 & CFG 7 & CFG 5 & CFG-distilled \\
\bottomrule
\end{tabular}
\end{table}

\subsection{\methodft{}: Adapter Training and Regression Targets}
\label{app:impl_ft}

\paragraph{Adapter.}
We insert rank-32 LoRA adapters ($\alpha=32$, no dropout, and zero-initialized
$B$ matrices) into the attention projections (query, key, value, and output)
and feed-forward layers up to and including the collapsed block. The adapter
is activated only in the student's first step; the remaining three steps use
the original student.

\paragraph{Targets and loss.}
Given a caption and noise sample, the student performs its first step with
the adapter enabled, and we extract the activation at the collapsed block.
The target is the teacher activation at the same block and step $\tau$.
The teacher starts from the same noise and caption, is advanced for
$\tau-1$ steps using the corresponding guidance, and is read from its
conditional branch at step $\tau$. We minimize plain MSE between the student
and teacher activations. For SDXL and Wan~2.1, the loss uses all generated
sample tokens. For SD3 and MiniMax-H3, which jointly process prompt and
sample tokens, we compute the loss only over image tokens and video tokens,
respectively. No additional loss terms are used.

\paragraph{Optimization.}
We train on the first 100 DiffusionDB captions
\citep{wang2023diffusiondb}, using 16 noise samples per caption for 1,600
caption--noise pairs, disjoint from the evaluation prompts. Each step uses
4 captions and 8 associated noises per caption, giving a batch of 32 pairs,
for 200 optimization steps. We use AdamW with learning rate $10^{-4}$,
weight decay 0.01, gradient clipping at 10, and a constant learning rate.
The forward pass uses bf16, while LoRA parameters are maintained in fp32.
We retain the final checkpoint and use the same recipe for all four pairs.

\subsection{\methodtf{}: Direction Estimation and Injection}
\label{app:impl_tf}

For each evaluation prompt, we run the teacher on $N=25$ noise samples,
independent of the evaluation noise bank, using the conditional branch.
We record the activation at step $\tau$ (7 for SDXL, 9 for SD3, 8 for
Wan~2.1, and 16 for MiniMax-H3), then compute and store the top $K=8$
directions $u_k$ and spreads $s_k$ from the Gram matrix following
App.~\ref{app:steering}.

During inference, each sample $i$ independently draws
$\alpha_{i,k}\sim\mathcal{U}[-2,2]$ and adds
$\sum_k\alpha_{i,k}s_k u_k$ to the student's activation at the collapsed
block. The perturbation is applied only during the first student step.   %

\section{Design Choices of \method{}}
\label{app:design}
\suppressfloats[t]

This section examines the key design choices underlying \method{} and establishes the configuration used in our final method. We organize the study around the two components of \method{}: \methodft{}, which determines how the student is regressed toward the teacher, and \methodtf{}, which determines how diversity directions are injected into the student. For \methodft{}, we systematically study where and when the regression is performed (Tables~\ref{tab:layer} and~\ref{tab:step}), which teacher step provides the regression target (Table~\ref{tab:kstep}), how teacher guidance is applied during regression (Table~\ref{tab:cfg}), when the learned adapter is activated at inference (Table~\ref{tab:gate}), and which blocks carry the LoRA adapter (Table~\ref{tab:placement}). For \methodtf{}, we study how many diversity directions are injected (Table~\ref{tab:ndir}) and which source provides these directions (Table~\ref{tab:randdir}). Each experiment changes one design choice while keeping the remaining components fixed at their defaults. All ablations are conducted on SD3. Each \methodft{} configuration (Tables~\ref{tab:layer}--\ref{tab:placement}) is fully retrained using the training recipe described in Sec.~\ref{sec:experiments_settings}; the \methodtf{} ablations (Tables~\ref{tab:ndir} and~\ref{tab:randdir}) involve no training and only change how the directions are injected at inference. We evaluate each configuration on the same 50 prompts, using 100 noise samples per prompt, for a total of 5,000 generated samples per configuration. This is a smaller evaluation set than the one behind Tab.~\ref{tab:main_results}, so the teacher and student values here differ from those reported there (for instance the student's DINO is $46.9$ rather than $48.4$); the rows of a table are comparable with each other, not across tables. We use the same evaluation metrics and protocol as in the main experiments; higher values indicate better performance for all reported metrics. The teacher is included as a reference and is not ranked, while the default configuration is highlighted throughout the tables.

\subsection{Which Block to Regress}
\label{app:layer}

\begin{table}[H]
\centering
\caption{\textbf{Regressed block.} We regress \methodft{} at different SD3 blocks, or at all 24 blocks at once (``All blocks''). Block 12, the collapsed block, is used as the default configuration. The teacher is shown only as a reference.}
\label{tab:layer}
\footnotesize
\setlength{\tabcolsep}{4pt}
\renewcommand{\arraystretch}{1.18}
\begin{adjustbox}{max width=\linewidth}
\begin{tabular}{@{}lccccccc@{}}
\toprule
& \multicolumn{4}{c}{\textbf{Diversity}} &
  \multicolumn{3}{c}{\textbf{Quality \& Alignment}} \\
\cmidrule(lr){2-5}\cmidrule(l){6-8}
\textbf{Setting}
& DINO$\uparrow$
& DreamSim$\uparrow$
& TPIPS$\uparrow$
& Vendi$\uparrow$
& CLIPScore$\uparrow$
& HPSv2$\uparrow$
& ImageReward$\uparrow$ \\
\midrule

\tg{Teacher}
& \tg{61.0} & \tg{32.9} & \tg{30.2} & \tg{6.26}
& \tg{31.8} & \tg{28.6} & \tg{0.885} \\

\addlinespace[2pt]
\midrule
\addlinespace[2pt]

Student
& 46.9 & 19.7 & 20.1 & 3.89
& \underline{31.9} & \textbf{29.3} & \textbf{0.955} \\

Block 8
& 48.9 & 21.9 & 21.6 & 4.33
& \underline{31.9} & 28.3 & 0.921 \\

Block 10
& 49.4 & 22.7 & 21.7 & 4.47
& \textbf{32.0} & 28.1 & 0.913 \\

\rowcolor{AblPick}
\textbf{Block 12}~\textit{(default)}
& 51.6 & 23.3 & 22.4 & 4.72
& \textbf{32.0} & \underline{28.4} & \underline{0.934} \\

Block 16
& 55.4 & 25.1 & 24.2 & 5.53
& 31.8 & 28.2 & 0.839 \\

Block 20
& 55.6 & 26.9 & 25.4 & 5.46
& 31.8 & 27.8 & 0.836 \\

Block 23
& \underline{57.4} & \underline{27.8} & \underline{26.2} & \underline{5.97}
& 31.8 & 27.8 & 0.831 \\

All blocks
& \textbf{57.8} & \textbf{28.5} & \textbf{26.6} & \textbf{6.05}
& 31.7 & 27.7 & 0.801 \\

\bottomrule
\end{tabular}
\end{adjustbox}
\end{table}

Regressing before the collapsed block provides only modest diversity gains over the student, with little change in quality (Table~\ref{tab:layer}). Regressing at later blocks recovers progressively more diversity, but quality degrades substantially, particularly in ImageReward. In contrast, the collapsed Block 12 provides a substantial diversity improvement while retaining the strongest quality among the regressed-block configurations: it achieves the highest HPSv2 and ImageReward among the ablated blocks while matching the best CLIPScore. Regressing at all blocks at once behaves like the latest blocks: it gives the highest diversity of the table but the lowest HPSv2 and ImageReward.

\subsection{Which Student Step and Which Teacher Step to Regress}
\label{app:step}

\begin{table}[H]
\centering
\caption{\textbf{Student step of the regression.} \methodft{} is regressed at different steps of the student's 4-step schedule, or at multiple steps. The adapter is active at the same steps at inference.}
\label{tab:step}
\footnotesize
\setlength{\tabcolsep}{4pt}
\renewcommand{\arraystretch}{1.18}
\begin{adjustbox}{max width=\linewidth}
\begin{tabular}{@{}lccccccc@{}}
\toprule
& \multicolumn{4}{c}{\textbf{Diversity}} &
  \multicolumn{3}{c}{\textbf{Quality \& Alignment}} \\
\cmidrule(lr){2-5}\cmidrule(l){6-8}
\textbf{Setting}
& DINO$\uparrow$
& DreamSim$\uparrow$
& TPIPS$\uparrow$
& Vendi$\uparrow$
& CLIPScore$\uparrow$
& HPSv2$\uparrow$
& ImageReward$\uparrow$ \\
\midrule

\tg{Teacher}
& \tg{61.0} & \tg{32.9} & \tg{30.2} & \tg{6.26}
& \tg{31.8} & \tg{28.6} & \tg{0.885} \\

\addlinespace[2pt]
\midrule
\addlinespace[2pt]

Student
& 46.9 & 19.7 & 20.1 & 3.89
& \underline{31.9} & \textbf{29.3} & \textbf{0.955} \\

\rowcolor{AblPick}
\textbf{Step 1}~\textit{(default)}
& \textbf{51.6} & 23.3 & 22.4 & 4.72
& \textbf{32.0} & \underline{28.4} & 0.934 \\

Step 2
& 48.2 & 23.0 & 21.2 & 5.13
& 31.2 & 25.8 & 0.477 \\

Step 3
& 47.7 & 21.7 & 20.0 & 4.60
& \underline{31.9} & 26.0 & 0.745 \\

Step 4
& 46.9 & 19.7 & 20.1 & 3.89
& \underline{31.9} & \textbf{29.3} & \underline{0.954} \\

Steps 1--2
& \underline{51.2} & \underline{26.3} & \textbf{23.2} & \underline{6.44}
& 31.0 & 24.9 & 0.299 \\

All four steps
& 50.0 & \textbf{27.7} & \underline{22.8} & \textbf{7.66}
& 29.4 & 20.0 & $-0.564$ \\

\bottomrule
\end{tabular}
\end{adjustbox}
\end{table}

Among the single-step configurations (Table~\ref{tab:step}), Step 1 provides the highest DINO diversity while retaining substantially better quality than Steps 2 and 3; Step 4 is effectively unchanged from the student. Activating the regression at multiple steps increases several diversity measures, particularly Vendi and DreamSim, but this comes with a pronounced degradation in CLIPScore, HPSv2, and ImageReward. Thus, Step 1 provides the strongest DINO diversity among the individual steps while avoiding the severe quality loss associated with multi-step regression.

\begin{table}[H]
\centering
\caption{\textbf{Teacher step $\tau$ of the regression target.} \methodft{} is regressed toward the teacher's activation at step $\tau$ of its 28-step schedule.}
\label{tab:kstep}
\footnotesize
\setlength{\tabcolsep}{4pt}
\renewcommand{\arraystretch}{1.18}
\begin{adjustbox}{max width=\linewidth}
\begin{tabular}{@{}lccccccc@{}}
\toprule
& \multicolumn{4}{c}{\textbf{Diversity}} &
  \multicolumn{3}{c}{\textbf{Quality \& Alignment}} \\
\cmidrule(lr){2-5}\cmidrule(l){6-8}
\textbf{Setting}
& DINO$\uparrow$
& DreamSim$\uparrow$
& TPIPS$\uparrow$
& Vendi$\uparrow$
& CLIPScore$\uparrow$
& HPSv2$\uparrow$
& ImageReward$\uparrow$ \\
\midrule

\tg{Teacher}
& \tg{61.0} & \tg{32.9} & \tg{30.2} & \tg{6.26}
& \tg{31.8} & \tg{28.6} & \tg{0.885} \\

\addlinespace[2pt]
\midrule
\addlinespace[2pt]

Student
& 46.9 & 19.7 & 20.1 & 3.89
& \underline{31.9} & \textbf{29.3} & \textbf{0.955} \\

$\tau{=}1$
& \underline{54.6} & \underline{26.5} & \textbf{24.8} & \underline{5.75}
& 31.5 & 26.3 & 0.642 \\

$\tau{=}3$
& 52.8 & 24.2 & 23.6 & 4.87
& \underline{31.9} & 28.6 & 0.906 \\

\rowcolor{AblPick}
\textbf{$\tau{=}9$}~\textit{(default)}
& 51.6 & 23.3 & 22.4 & 4.72
& \textbf{32.0} & 28.4 & 0.934 \\

$\tau{=}18$
& 52.6 & 24.1 & 22.6 & 4.90
& \textbf{32.0} & 27.6 & 0.820 \\

$\tau{=}28$
& \textbf{58.7} & \textbf{28.8} & \underline{24.7} & \textbf{7.16}
& 31.1 & 24.0 & 0.328 \\

\bottomrule
\end{tabular}
\end{adjustbox}
\end{table}

The teacher step reveals a clear trade-off between diversity and quality (Table~\ref{tab:kstep}). Early targets such as $\tau{=}1$ substantially improve diversity over the student, but lead to lower HPSv2 and ImageReward. At the other extreme, $\tau{=}28$ achieves the strongest diversity, with the highest DINO, DreamSim, and Vendi scores, but incurs a substantial drop in quality and alignment. Among these settings, $\tau{=}9$ provides a strong overall balance and is therefore used as the default.

\subsection{Teacher Guidance During Regression}
\label{app:cfg}

\begin{table}[H]
\centering
\caption{\textbf{Teacher target.} The regression target is obtained from the teacher's conditional branch, from the teacher rolled under classifier-free guidance at scale $w$, or from its unconditional branch. The released scale $w{=}7$ is the default.}
\label{tab:cfg}
\footnotesize
\setlength{\tabcolsep}{4pt}
\renewcommand{\arraystretch}{1.18}
\begin{adjustbox}{max width=\linewidth}
\begin{tabular}{@{}lccccccc@{}}
\toprule
& \multicolumn{4}{c}{\textbf{Diversity}} &
  \multicolumn{3}{c}{\textbf{Quality \& Alignment}} \\
\cmidrule(lr){2-5}\cmidrule(l){6-8}
\textbf{Setting}
& DINO$\uparrow$
& DreamSim$\uparrow$
& TPIPS$\uparrow$
& Vendi$\uparrow$
& CLIPScore$\uparrow$
& HPSv2$\uparrow$
& ImageReward$\uparrow$ \\
\midrule

\tg{Teacher}
& \tg{61.0} & \tg{32.9} & \tg{30.2} & \tg{6.26}
& \tg{31.8} & \tg{28.6} & \tg{0.885} \\

\addlinespace[2pt]
\midrule
\addlinespace[2pt]

Student
& 46.9 & 19.7 & 20.1 & 3.89
& \underline{31.9} & \textbf{29.3} & \textbf{0.955} \\

Conditional, no guidance
& \textbf{58.6} & \textbf{27.6} & \textbf{26.4} & \textbf{5.82}
& 31.8 & 27.9 & 0.807 \\

Guidance $w{=}2$
& \underline{56.1} & \underline{25.6} & \underline{24.8} & \underline{5.33}
& \underline{31.9} & 28.3 & 0.882 \\

Guidance $w{=}3.5$
& 53.8 & 24.3 & 23.5 & 4.97
& \underline{31.9} & \underline{28.5} & 0.911 \\

Guidance $w{=}5$
& 52.6 & 23.7 & 23.0 & 4.84
& \textbf{32.0} & \underline{28.5} & 0.921 \\

\rowcolor{AblPick}
\textbf{Guidance $w{=}7$}~\textit{(default)}
& 51.6 & 23.3 & 22.4 & 4.72
& \textbf{32.0} & 28.4 & \underline{0.934} \\

Unconditional
& 51.9 & 23.7 & 22.7 & 4.86
& \textbf{32.0} & 28.4 & 0.925 \\

\bottomrule
\end{tabular}
\end{adjustbox}
\end{table}

Teacher guidance controls the trade-off between diversity and quality (Table~\ref{tab:cfg}). Removing guidance produces substantially higher diversity but lowers ImageReward, while increasing the guidance scale progressively reduces diversity as the target becomes more aligned with the teacher's guided trajectory. The default $w{=}7$ provides lower diversity than weakly guided targets but achieves the highest ImageReward among the guided configurations while maintaining the best CLIPScore.

\subsection{Which Steps to Open the LoRA at Inference}
\label{app:gate}

\begin{table}[H]
\centering
\caption{\textbf{Adapter schedule at inference.} The adapter is activated at the indicated steps; ``never'' corresponds to the unmodified student. The default activates the adapter only at Step 1.}
\label{tab:gate}
\footnotesize
\setlength{\tabcolsep}{4pt}
\renewcommand{\arraystretch}{1.18}
\begin{adjustbox}{max width=\linewidth}
\begin{tabular}{@{}lccccccc@{}}
\toprule
& \multicolumn{4}{c}{\textbf{Diversity}} &
  \multicolumn{3}{c}{\textbf{Quality \& Alignment}} \\
\cmidrule(lr){2-5}\cmidrule(l){6-8}
\textbf{Setting}
& DINO$\uparrow$
& DreamSim$\uparrow$
& TPIPS$\uparrow$
& Vendi$\uparrow$
& CLIPScore$\uparrow$
& HPSv2$\uparrow$
& ImageReward$\uparrow$ \\
\midrule

\tg{Teacher}
& \tg{61.0} & \tg{32.9} & \tg{30.2} & \tg{6.26}
& \tg{31.8} & \tg{28.6} & \tg{0.885} \\

\addlinespace[2pt]
\midrule
\addlinespace[2pt]

\rowcolor{AblPick}
\textbf{Step 1 only}~\textit{(default)}
& \underline{51.6} & 23.3 & 22.4 & 4.72
& \textbf{32.0} & \underline{28.4} & \underline{0.934} \\

Step 2 only
& 46.9 & 21.2 & 20.8 & 4.31
& 31.7 & 27.9 & 0.788 \\

Steps 1--2
& \textbf{51.7} & \underline{25.8} & \underline{23.4} & \underline{5.91}
& 31.6 & 26.4 & 0.607 \\

All four steps
& \textbf{51.7} & \textbf{27.0} & \textbf{23.8} & \textbf{7.02}
& 31.3 & 25.1 & 0.380 \\

Never (student)
& 46.9 & 19.7 & 20.1 & 3.89
& \underline{31.9} & \textbf{29.3} & \textbf{0.955} \\

\bottomrule
\end{tabular}
\end{adjustbox}
\end{table}

Activating the adapter only at Step 1 provides a substantial gain in DINO diversity while largely preserving quality and alignment (Table~\ref{tab:gate}). Activating it at later steps alone does not improve diversity over Step 1 and can reduce quality, particularly ImageReward. Applying the adapter across multiple steps further increases several diversity metrics, most notably Vendi, but consistently degrades HPSv2 and ImageReward. These results support the default Step-1-only schedule as a balanced trade-off between diversity and quality.

\subsection{Which Blocks Carry the LoRA}
\label{app:placement}

\begin{table}[H]
\centering
\caption{\textbf{Adapter placement.} We vary the blocks and modules carrying the LoRA adapter. ``attn'' denotes attention projections, ``FFN'' denotes feed-forward layers, and $\ell{=}12$ denotes the collapsed block. The default covers all blocks up to $\ell$ in both attention and FFN layers.}
\label{tab:placement}
\footnotesize
\setlength{\tabcolsep}{4pt}
\renewcommand{\arraystretch}{1.18}
\begin{adjustbox}{max width=\linewidth}
\begin{tabular}{@{}lccccccc@{}}
\toprule
& \multicolumn{4}{c}{\textbf{Diversity}} &
  \multicolumn{3}{c}{\textbf{Quality \& Alignment}} \\
\cmidrule(lr){2-5}\cmidrule(l){6-8}
\textbf{Setting}
& DINO$\uparrow$
& DreamSim$\uparrow$
& TPIPS$\uparrow$
& Vendi$\uparrow$
& CLIPScore$\uparrow$
& HPSv2$\uparrow$
& ImageReward$\uparrow$ \\
\midrule

\tg{Teacher}
& \tg{61.0} & \tg{32.9} & \tg{30.2} & \tg{6.26}
& \tg{31.8} & \tg{28.6} & \tg{0.885} \\

\addlinespace[2pt]
\midrule
\addlinespace[2pt]

Student
& 46.9 & 19.7 & 20.1 & 3.89
& \underline{31.9} & \textbf{29.3} & \textbf{0.955} \\

\rowcolor{AblPick}
\textbf{Blocks $\leq \ell$, attn + FFN}~\textit{(default)}
& \textbf{51.6} & \textbf{23.3} & \textbf{22.4} & \textbf{4.72}
& \textbf{32.0} & 28.4 & 0.934 \\

Blocks $\leq \ell$, attn only
& \underline{50.5} & \textbf{23.3} & \textbf{22.4} & \underline{4.62}
& \textbf{32.0} & 28.4 & 0.923 \\

Blocks $6$--$12$
& 50.3 & \underline{22.7} & \underline{21.9} & 4.55
& \textbf{32.0} & 28.3 & 0.921 \\

Block $\ell$ only
& 47.8 & 20.8 & 20.3 & 4.16
& \textbf{32.0} & \underline{28.6} & \underline{0.935} \\

\bottomrule
\end{tabular}
\end{adjustbox}
\end{table}

The placement study (Table~\ref{tab:placement}) shows that adapter coverage primarily affects diversity, while quality and alignment remain comparatively stable across configurations. Expanding the adapter from the collapsed block alone to a broader range of preceding blocks generally increases diversity, with the full placement through the collapsed block achieving the highest DINO, DreamSim, TPIPS, and Vendi scores among the tested configurations.

\subsection{How Many Directions to Inject}
\label{app:ndir}

\begin{table}[H]
\centering
\caption{\textbf{Number of injected directions $K$.} \methodtf{} injects the top $K$ teacher directions into the student; $K{=}0$ corresponds to the unmodified student.}
\label{tab:ndir}
\footnotesize
\setlength{\tabcolsep}{4pt}
\renewcommand{\arraystretch}{1.18}
\begin{adjustbox}{max width=\linewidth}
\begin{tabular}{@{}lccccccc@{}}
\toprule
& \multicolumn{4}{c}{\textbf{Diversity}} &
  \multicolumn{3}{c}{\textbf{Quality \& Alignment}} \\
\cmidrule(lr){2-5}\cmidrule(l){6-8}
\textbf{Setting}
& DINO$\uparrow$
& DreamSim$\uparrow$
& TPIPS$\uparrow$
& Vendi$\uparrow$
& CLIPScore$\uparrow$
& HPSv2$\uparrow$
& ImageReward$\uparrow$ \\
\midrule

\tg{Teacher}
& \tg{61.0} & \tg{32.9} & \tg{30.2} & \tg{6.26}
& \tg{31.8} & \tg{28.6} & \tg{0.885} \\

\addlinespace[2pt]
\midrule
\addlinespace[2pt]

Student ($K{=}0$)
& 46.9 & 19.7 & 20.1 & 3.89
& \textbf{31.9} & \textbf{29.3} & \textbf{0.955} \\

$K{=}1$
& 48.2 & 20.7 & 20.8 & 4.12
& \textbf{31.9} & \underline{29.2} & \underline{0.941} \\

$K{=}2$
& 49.1 & 21.5 & 21.5 & 4.31
& \underline{31.8} & 29.1 & 0.929 \\

$K{=}4$
& 50.6 & 22.6 & 22.5 & 4.67
& 31.7 & 28.9 & 0.907 \\

\rowcolor{AblPick}
\textbf{$K{=}8$}~\textit{(default)}
& 53.0 & 24.5 & 24.1 & 5.18
& 31.6 & 28.8 & 0.878 \\

$K{=}16$
& \underline{56.3} & \underline{27.3} & \underline{26.4} & \underline{5.85}
& 31.3 & 28.6 & 0.811 \\

$K{=}24$
& \textbf{58.4} & \textbf{29.8} & \textbf{28.2} & \textbf{6.57}
& 31.1 & 28.5 & 0.761 \\

\bottomrule
\end{tabular}
\end{adjustbox}
\end{table}

Increasing $K$ consistently improves diversity across all four metrics (Table~\ref{tab:ndir}), but this gain comes with a gradual decline in image quality and alignment. The trade-off becomes increasingly pronounced as more directions are injected, with the largest direction sets showing noticeable degradation in quality. We therefore use $K{=}8$ as the default, as it provides a substantial diversity improvement while preserving image quality and alignment more effectively than larger direction sets.

\subsection{Whose Directions: Teacher, Student or Random}
\label{app:randdir}

\begin{table}[H]
\centering
\caption{\textbf{Source of the injected directions.} We inject $K{=}8$ directions obtained from the teacher, the student itself, or random directions with the teacher's spreads, using the same injected energy.}
\label{tab:randdir}
\footnotesize
\setlength{\tabcolsep}{3.2pt}
\renewcommand{\arraystretch}{1.18}
\begin{adjustbox}{max width=\linewidth}
\begin{tabular}{@{}lccccccccccc@{}}
\toprule
& \multicolumn{8}{c}{\textbf{Diversity}} &
  \multicolumn{3}{c}{\textbf{Quality \& Alignment}} \\
\cmidrule(lr){2-9}\cmidrule(l){10-12}
\textbf{Setting}
& DINO$\uparrow$
& DreamSim$\uparrow$
& Vendi$\uparrow$
& TPIPS$_{\text{ove}}\uparrow$
& TPIPS$_{\text{subj}}\uparrow$
& TPIPS$_{\text{bg}}\uparrow$
& TPIPS$_{\text{cam}}\uparrow$
& TPIPS$_{\text{light}}\uparrow$
& CLIPScore$\uparrow$
& HPSv2$\uparrow$
& ImageReward$\uparrow$ \\
\midrule

\tg{Teacher}
& \tg{61.0} & \tg{32.9} & \tg{6.26} & \tg{30.2}
& \tg{20.4} & \tg{20.9} & \tg{20.3} & \tg{16.5}
& \tg{31.8} & \tg{28.6} & \tg{0.885} \\

\addlinespace[2pt]
\midrule
\addlinespace[2pt]

Student
& 46.9 & 19.7 & 3.89 & 20.1
& 13.9 & 12.9 & 11.9 & 10.7
& \textbf{31.9} & \underline{29.3} & \textbf{0.955} \\

\rowcolor{AblPick}
\textbf{Teacher's directions}~\textit{(default)}
& \underline{53.0} & \underline{24.5} & \underline{5.18} & \underline{24.1}
& \underline{16.5} & \underline{17.9} & \underline{14.3} & \underline{13.6}
& 31.6 & 28.8 & 0.878 \\

Student's own directions
& 49.5 & 20.7 & 4.33 & 21.0
& 14.5 & 14.5 & 12.4 & 11.3
& \underline{31.8} & \textbf{29.4} & \underline{0.954} \\

Random directions
& \textbf{57.8} & \textbf{25.3} & \textbf{6.16} & \textbf{24.5}
& \textbf{17.6} & \textbf{20.8} & \textbf{15.0} & \textbf{14.0}
& 30.8 & 28.7 & 0.719 \\

\bottomrule
\end{tabular}
\end{adjustbox}
\end{table}

The source of the injected directions determines both the magnitude and the character of the resulting diversity (Table~\ref{tab:randdir}). The student's own directions provide only modest gains over the unmodified student across both aggregate and aspect-conditioned diversity, suggesting that they contribute limited additional variation beyond what the student already represents. Random directions produce the largest diversity gains, with improvements spanning subject, background, camera, and lighting, but this broader variation comes with a substantially larger loss in image quality and alignment. In contrast, teacher directions achieve strong diversity gains while maintaining substantially better alignment, and their aspect-conditioned gains are more balanced across the different visual factors rather than being dominated by a single source of variation. Together, these results indicate that teacher directions provide a more controlled form of diversity injection: they introduce meaningful variation while preserving the semantic and perceptual characteristics of the generated images. We therefore use teacher directions as the default source.   %
\section{Evaluation Protocol and Metrics}
\label{app:protocol}

This section describes the evaluation prompts, noise construction, subsets, and metrics used throughout our experiments. Comparisons are paired: all methods are evaluated on the same prompts and the same initial noise samples.

\subsection{Prompts, Seeds, and Evaluation Subsets}
\label{app:protocol_prompts}

For images, we use the 553 GenEval prompts and generate 100 samples per prompt at $1024{\times}1024$ resolution.
For video, we use 300 prompts selected from VBench with a fixed quota for each dimension and generate 10 clips per prompt.
Wan~2.1 generates 81-frame clips at $832{\times}480$ and 16~fps, while MiniMax-H3 generates 124-frame clips at
$1344{\times}768$ and 24~fps.

For each backbone, we use a fixed stored noise bank, so paired methods receive identical initial noise.
The coefficients $\alpha$ used by \methodtf{} are also generated deterministically, ensuring that \ourstest{} (the student carrying the directions, with no adapter) and \ours{} use identical coefficients.
The noises used for direction estimation and the DiffusionDB captions used to train \methodft{} are disjoint from all evaluation prompts and noises.

\subsection{Diversity Metrics}
\label{app:protocol_metrics}

Diversity metrics are computed within each prompt and then averaged across prompts.
Except for Vendi, which evaluates the full sample set jointly, and LPIPS, which is a learned perceptual
distance, the remaining metrics compare pairs of samples from the same prompt using one minus the cosine
similarity between their embeddings.
Thus, higher values indicate greater diversity.

\paragraph{DINO.}
For images, we use DINOv2-B patch tokens and compare corresponding spatial locations, averaging their cosine distance.
This captures differences in layout, pose, and content.
For video, we sample 16 frames throughout each clip, including the first and last, and average the corresponding frame distances.

\paragraph{DreamSim.}
DreamSim combines DINO, CLIP, and OpenCLIP features and is fine-tuned to human judgments of visual similarity.
It captures perceptual differences in layout, pose, color, and object appearance.

\paragraph{LPIPS.}
LPIPS measures distance between deep VGG features and is sensitive to differences in texture, color, and local structure, including variations that preserve semantic content.

\paragraph{TPIPS.}
TPIPS conditions the visual embedding on the aspect being compared.
We use ``Represent the similarity of the image.'' for the overall score and
``Represent the similarity of the image based on X,'' for the main subject, background scene, lighting and color, and camera viewpoint and framing.
$\mathrm{TPIPS}_{\mathrm{avg}}$ is the mean of the overall and four aspect-specific scores.
For video, we sample 8 frames and compute the same comparison frame by frame.

\paragraph{Vendi.}
Vendi measures set-level diversity rather than individual pairwise differences.
Given the normalized similarity matrix $K$, its score is the exponential of the entropy of the eigenvalue distribution of $K$, reflecting the effective number of distinct samples generated for a prompt.
For images, we use DINOv2-B CLS features; for video, we use ViCLIP embeddings.

\subsection{Video-Specific Diversity}
\label{app:protocol_metrics_video}

Video diversity metrics either compare corresponding frames or represent each clip with a single video embedding.

\paragraph{V-JEPA~2.}
We embed 64 frames from each clip using V-JEPA~2 ViT-L and average the resulting tokens.
Because the model processes the frames jointly, the metric captures differences in both appearance and motion.

\paragraph{ViCLIP.}
ViCLIP is the video--text model used by VBench.
We use 8 frames per clip and compute the distance between clip-level embeddings.
ViCLIP$_{\mathrm{div}}$ therefore captures differences in overall content, while the ViCLIP Vendi score measures the effective number of distinct clips per prompt.

\paragraph{Camera trajectory.}
We estimate camera trajectories with ViPE and express them relative to the first camera position.
Each trajectory is resampled to 32 points and normalized by its path length, removing absolute-scale differences.
CamTraj is the mean pointwise distance between two normalized trajectories, while CamEnd is the distance between their endpoints. Both are reported after the per-prompt normalization described above: the ``Cam'' column of Tab.~\ref{tab:main_results} is the normalized CamTraj and the ``CamEnd'' column of Tab.~\ref{tab:supp_tpips} is the normalized CamEnd, so the two columns are not on the same scale and should not be compared with each other.

\subsection{Quality, Alignment, and Reporting}
\label{app:protocol_quality}

\paragraph{Images.}
CLIP-S measures prompt alignment using the cosine similarity between CLIP (ViT-B/32) image and text embeddings.
HPSv2 (version 2.1) and ImageReward are preference models trained on human judgments of generated images;
HPSv2 uses a CLIP ViT-H/14 backbone. 

\paragraph{Video.}
SubjC and BgC are VBench dimensions measuring subject and background consistency across frames using DINO and CLIP features, respectively.
ViCLIP-S measures prompt alignment using cosine similarity between the generated clip and prompt embeddings.

\paragraph{Reporting.}
Diversity metrics are computed within each prompt and averaged across prompts, while quality and alignment metrics are averaged over all generated samples.
All reported scores are multiplied by 100 except Vendi, ImageReward, and PickScore.
In Tab.~\ref{tab:main_results}, Teacher and Noise Optimization are reference methods and are not ranked.
Among the remaining methods, the best and second-best values are shown in bold and underlined, respectively.         %
\section{Baseline Reproductions}
\label{app:baselines}

None of the baselines provides checkpoints trained with our teacher--student pairs, so we reproduced each baseline ourselves.
All baselines use the same initial noise bank as the other methods and are evaluated on the full evaluation sets using the
protocol described in App.~\ref{app:protocol}.

\subsection{Image Baselines}
\label{app:baselines_image}

\paragraph{Noise Optimization.}
Noise Optimization \citep{harrington2026s} keeps the student fixed and optimizes the initial noise at inference time.
It optimizes groups of four noise samples using gradient ascent on a diversity objective based on DINOv2 features, together
with an HPSv2 objective to preserve image quality. We use the authors' code and, where applicable, their recommended
settings. We use 50 optimization iterations for both SDXL and SD3 to keep the evaluation computationally feasible.
For SDXL, the diversity objective is evaluated using DPP with 150 points and the HPSv2 objective uses a weight of 3.
For SD3, we use 100 DPP points and an HPSv2 weight of 5. Since Noise Optimization directly optimizes a
DINOv2-based diversity objective, we report it as an optimization-based reference in Tab.~\ref{tab:main_results}.

\paragraph{DP-DMD.}
DP-DMD \citep{wu2026diversitypreserved} distills a new 4-step student using DMD and regresses the student's first step toward the
teacher's latent representation at an earlier point in the teacher's denoising trajectory. This is intended to preserve
the teacher's diversity in the distilled student. The released checkpoint is distilled from SD3.5-Medium rather than
from our teacher, so we train new students for both SD3 and SDXL from scratch on DiffusionDB prompts. For SD3, we use
the authors' official implementation. For SDXL, we use our own implementation based on the same training procedure,
as the official implementation does not support SDXL.

\subsection{Video Baselines}
\label{app:baselines_video}

\paragraph{DFD.}
DFD \citep{chen2026dataforcing} post-trains the student with DMD while incorporating real videos into the training
process. In half of the updates, the real score is evaluated on a noised real video with the same caption instead of
on a generated sample, encouraging the student to remain closer to the real data distribution. For Wan~2.1, we use
the authors' implementation with 29k captioned real video clips. MiniMax-H3 is not supported by the released code,
so we adapt the method to its student architecture and train it using 2k real video clips. For both models, we first
warm up the fake score for 100 iterations and then train DFD for another 100 iterations. We also evaluate several
variants of this setup and use the best configuration for DFD in the baseline comparison.        %
\section{Additional Quantitative Results}
\label{app:results}

We provide three additional views of the results in Tab.~\ref{tab:main_results}:
paired comparisons with the student, per-aspect TPIPS scores, and diversity
measured separately on foreground and background regions.

\subsection{Paired Comparison with the Student}
\label{app:results_paired}

Because all methods are evaluated with the same prompts and initial noises,
we can compare each method with the student on a prompt-by-prompt basis.
As shown in Tab.~\ref{tab:supp_paired}, \method{} improves diversity on the
vast majority of prompts across all backbones, while alignment improves on
a smaller but substantial fraction of prompts. Thus, the increased diversity
is generally accompanied by only a limited loss of alignment.

\begin{table}[h]
\centering
\caption{\textbf{Paired comparison with the student.} Percentage of prompts on which each method scores higher than the student using the same prompt and initial noise. We report DINO and TPIPS diversity, and CLIPScore for images or ViCLIP-S for video. All DINO differences are significant ($p<0.001$, two-sided sign test). \textbf{Bold} and \underline{underline} mark the highest and second-highest value below the dashed line; the teacher and Noise Optimization are references and are not ranked. Two methods can share a win rate while differing in how much they win by: on SD3, \ourstest{} and \ours{} both beat the student on $100\%$ of the prompts in DINO, by $+5.5$ and $+11.7$ points respectively.}
\label{tab:supp_paired}
\tabstyle
\begin{tabularx}{\linewidth}{@{}l *{3}{>{\centering\arraybackslash}X}@{}}
\toprule
\textbf{Method} & \textbf{DINO} & \textbf{TPIPS} & \textbf{CLIPScore} \\
\midrule
\rowcolor{BlockGray}
\multicolumn{4}{@{}l}{{\scriptsize\bfseries\itshape SDXL}} \\
\tg{Teacher} & \tg{96} & \tg{98} & \tg{48} \\
\tg{Noise Opt.} & \tg{100} & \tg{100} & \tg{25} \\
\addlinespace[1pt]\cdashline{1-4}[1.2pt/1.2pt]\addlinespace[2pt]
DP-DMD & 37 & 24 & 35 \\
\addlinespace[2pt]
\rowcolor{OursGreen} \ourstest{} & \underline{88} & \underline{85} & \textbf{44} \\
\rowcolor{OursGreen} \ourstrain{} & \textbf{95} & \textbf{97} & \underline{41} \\
\rowcolor{OursGreen} \ours{} & \textbf{95} & \textbf{97} & 38 \\
\midrule
\rowcolor{BlockGray}
\multicolumn{4}{@{}l}{{\scriptsize\bfseries\itshape SD3}} \\
\tg{Teacher} & \tg{99} & \tg{98} & \tg{47} \\
\tg{Noise Opt.} & \tg{100} & \tg{100} & \tg{28} \\
\addlinespace[1pt]\cdashline{1-4}[1.2pt/1.2pt]\addlinespace[2pt]
DP-DMD & \underline{97} & 81 & 21 \\
\addlinespace[2pt]
\rowcolor{OursGreen} \ourstest{} & \textbf{100} & \textbf{99} & \underline{31} \\
\rowcolor{OursGreen} \ourstrain{} & 94 & \underline{83} & \textbf{58} \\
\rowcolor{OursGreen} \ours{} & \textbf{100} & \textbf{99} & \underline{31} \\
\bottomrule
\end{tabularx}

\vspace{6pt}

\begin{tabularx}{\linewidth}{@{}l *{3}{>{\centering\arraybackslash}X}@{}}
\toprule
\textbf{Method} & \textbf{DINO} & \textbf{TPIPS} & \textbf{ViCLIP-S} \\
\midrule
\rowcolor{BlockGray}
\multicolumn{4}{@{}l}{{\scriptsize\bfseries\itshape Wan 2.1}} \\
\tg{Teacher} & \tg{93} & \tg{94} & \tg{45} \\
\addlinespace[1pt]\cdashline{1-4}[1.2pt/1.2pt]\addlinespace[2pt]
DFD & 26 & 15 & \textbf{59} \\
\addlinespace[2pt]
\rowcolor{OursGreen} \ourstest{} & 91 & \textbf{98} & \underline{50} \\
\rowcolor{OursGreen} \ourstrain{} & \underline{93} & \underline{91} & 40 \\
\rowcolor{OursGreen} \ours{} & \textbf{98} & \textbf{98} & 36 \\
\midrule
\rowcolor{BlockGray}
\multicolumn{4}{@{}l}{{\scriptsize\bfseries\itshape MiniMax-H3}} \\
\tg{Teacher} & \tg{87} & \tg{86} & \tg{61} \\
\addlinespace[1pt]\cdashline{1-4}[1.2pt/1.2pt]\addlinespace[2pt]
DFD & 40 & 40 & \textbf{58} \\
\addlinespace[2pt]
\rowcolor{OursGreen} \ourstest{} & \textbf{92} & \textbf{97} & 47 \\
\rowcolor{OursGreen} \ourstrain{} & 75 & 77 & \underline{53} \\
\rowcolor{OursGreen} \ours{} & \underline{88} & \underline{91} & 47 \\
\bottomrule
\end{tabularx}
\end{table}

\subsection{Per-Aspect TPIPS}
\label{app:results_tpips}

Tab.~\ref{tab:supp_tpips} breaks down TPIPS into its individual aspects.
Across the image and video backbones, \method{} improves diversity across
multiple aspects, including subject, background, lighting, and camera
viewpoint, indicating that the improvement is not concentrated in a single
source of variation.

\begin{table}[h]
\centering
\caption{\textbf{Per-aspect TPIPS.} TPIPS scores for the overall evaluation set and the four aspect-specific subsets. For video, we additionally report CamEnd and the ViCLIP Vendi score. The overall score is computed over all prompts, while the four aspect scores use the 100-prompt subset; CamEnd uses the 30-prompt subset. TPIPS and CamEnd are scaled by $100$; Vendi is unscaled. All metrics are higher-is-better. \textbf{Bold} and \underline{underline} mark the highest and second-highest value below the dashed line; the teacher and Noise Optimization are references and are not ranked.}
\label{tab:supp_tpips}
\tabstyledense%

\begin{tabularx}{\linewidth}{@{}l *{5}{>{\centering\arraybackslash}X}@{}}
\toprule
 & \multicolumn{5}{c}{\textbf{TPIPS}} \\
\cmidrule(l){2-6}
\textbf{Method} & \textbf{Overall} & \textbf{Subject} & \textbf{Backgr.} & \textbf{Lighting} & \textbf{Camera} \\
\midrule
\rowcolor{BlockGray}
\multicolumn{6}{@{}l}{{\scriptsize\bfseries\itshape SDXL}} \\
\tg{Teacher} & \tg{32.3} & \tg{23.5} & \tg{23.4} & \tg{17.1} & \tg{24.0} \\
\tg{Noise Opt.} & \tg{33.6} & \tg{23.8} & \tg{23.6} & \tg{20.3} & \tg{24.4} \\
\addlinespace[1pt]\cdashline{1-6}[1.2pt/1.2pt]\addlinespace[2pt]
Student & 25.5 & 17.5 & 18.5 & 15.4 & 18.2 \\
DP-DMD & 23.1 & 16.5 & 18.6 & 13.7 & 16.3 \\
\rowcolor{OursGreen} \ourstest{} & 26.2 & 18.0 & 19.2 & 15.9 & 18.6 \\
\rowcolor{OursGreen} \ourstrain{} & \underline{29.9} & \underline{21.4} & \underline{22.6} & \underline{17.3} & \underline{21.8} \\
\rowcolor{OursGreen} \ours{} & \textbf{30.3} & \textbf{21.7} & \textbf{23.1} & \textbf{17.5} & \textbf{22.2} \\
\midrule
\rowcolor{BlockGray}
\multicolumn{6}{@{}l}{{\scriptsize\bfseries\itshape SD3}} \\
\tg{Teacher} & \tg{27.7} & \tg{20.2} & \tg{21.1} & \tg{15.8} & \tg{20.9} \\
\tg{Noise Opt.} & \tg{25.5} & \tg{19.1} & \tg{17.1} & \tg{13.1} & \tg{16.9} \\
\addlinespace[1pt]\cdashline{1-6}[1.2pt/1.2pt]\addlinespace[2pt]
Student & 19.2 & 13.9 & 13.2 & 10.1 & 12.4 \\
DP-DMD & 22.4 & \underline{16.5} & \underline{18.5} & \underline{13.0} & \underline{16.6} \\
\rowcolor{OursGreen} \ourstest{} & \underline{22.8} & 16.4 & 18.1 & 12.9 & 14.7 \\
\rowcolor{OursGreen} \ourstrain{} & 21.6 & 15.8 & 14.5 & 10.6 & 14.2 \\
\rowcolor{OursGreen} \ours{} & \textbf{26.9} & \textbf{19.5} & \textbf{21.8} & \textbf{15.4} & \textbf{18.0} \\
\bottomrule
\end{tabularx}

\vspace{6pt}

\begin{tabularx}{\linewidth}{@{}l *{7}{>{\centering\arraybackslash}X}@{}}
\toprule
 & \multicolumn{5}{c}{\textbf{TPIPS}} & & \\
\cmidrule(lr){2-6}
\textbf{Method} & \textbf{Overall} & \textbf{Subject} & \textbf{Backgr.} & \textbf{Lighting} & \textbf{Camera} & \textbf{CamEnd} & \textbf{ViCLIP Vendi} \\
\midrule
\rowcolor{BlockGray}
\multicolumn{8}{@{}l}{{\scriptsize\bfseries\itshape Wan 2.1}} \\
\tg{Teacher} & \tg{41.4} & \tg{31.4} & \tg{29.7} & \tg{25.2} & \tg{28.9} & \tg{69.8} & \tg{3.72} \\
\addlinespace[1pt]\cdashline{1-8}[1.2pt/1.2pt]\addlinespace[2pt]
Student & 30.1 & 23.5 & 24.5 & 16.7 & 21.4 & 66.7 & 2.51 \\
DFD & 23.7 & 18.1 & 19.0 & 12.5 & 15.9 & 46.3 & 2.23 \\
\rowcolor{OursGreen} \ourstest{} & 33.6 & 25.3 & 27.1 & \underline{22.6} & 22.9 & 66.3 & 2.81 \\
\rowcolor{OursGreen} \ourstrain{} & \underline{38.2} & \underline{30.3} & \underline{29.4} & 19.7 & \underline{26.0} & \textbf{78.6} & \underline{3.63} \\
\rowcolor{OursGreen} \ours{} & \textbf{42.5} & \textbf{32.9} & \textbf{31.8} & \textbf{26.8} & \textbf{28.8} & \underline{78.1} & \textbf{4.11} \\
\midrule
\rowcolor{BlockGray}
\multicolumn{8}{@{}l}{{\scriptsize\bfseries\itshape MiniMax-H3}} \\
\tg{Teacher} & \tg{40.3} & \tg{32.1} & \tg{29.2} & \tg{20.3} & \tg{29.9} & \tg{70.4} & \tg{3.74} \\
\addlinespace[1pt]\cdashline{1-8}[1.2pt/1.2pt]\addlinespace[2pt]
Student & 30.2 & 25.1 & 23.5 & 15.1 & 21.3 & 47.6 & 2.97 \\
DFD & 29.5 & 24.1 & 22.7 & 14.4 & 20.5 & 46.9 & 2.88 \\
\rowcolor{OursGreen} \ourstest{} & 34.3 & 28.1 & \underline{26.8} & \underline{19.0} & 23.6 & 54.3 & \underline{3.48} \\
\rowcolor{OursGreen} \ourstrain{} & \underline{34.8} & \underline{28.3} & 25.9 & 16.7 & \underline{25.1} & \underline{61.7} & 3.34 \\
\rowcolor{OursGreen} \ours{} & \textbf{39.1} & \textbf{31.9} & \textbf{29.5} & \textbf{20.8} & \textbf{28.0} & \textbf{66.7} & \textbf{3.98} \\
\bottomrule
\end{tabularx}
\end{table}

\subsection{Foreground and Background Diversity}
\label{app:results_fgbg}

Tab.~\ref{tab:fgbg} separates each sample into its foreground, segmented
with SAM~3, and the remaining background. \method{} improves diversity in
both regions across all backbones, indicating that the recovered variation
is not restricted to the background.

\begin{table}[H]
\caption{\textbf{Foreground and background diversity.} Diversity of the SAM~3 foreground and of the background, measured separately with DINO and CLIP embeddings, and of the foreground's mask geometry (Pose); mean pairwise distance over the samples of a prompt, averaged over prompts, $\times 100$. Higher is better. \textbf{Bold} and \underline{underline} mark the highest and second-highest value below the dashed line; the teacher and Noise Optimization are references and are not ranked.}
\label{tab:fgbg}
\centering
\tabstyle
\begin{tabularx}{\linewidth}{@{}l *{5}{>{\centering\arraybackslash}X}@{}}
\toprule
 & \multicolumn{2}{c}{\textbf{Foreground}} & \multicolumn{2}{c}{\textbf{Background}} & \\
\cmidrule(lr){2-3} \cmidrule(lr){4-5}
\textbf{Method} & \textbf{DINO} & \textbf{CLIP} & \textbf{DINO} & \textbf{CLIP} & \textbf{Pose} \\
\midrule
\rowcolor{BlockGray}
\multicolumn{6}{@{}l}{{\scriptsize\bfseries\itshape SDXL}} \\
\tg{Teacher} & \tg{39.0} & \tg{16.7} & \tg{50.1} & \tg{19.2} & \tg{21.3} \\
\tg{Noise Opt.} & \tg{41.8} & \tg{16.3} & \tg{52.1} & \tg{18.7} & \tg{19.2} \\
\addlinespace[1pt]\cdashline{1-6}[1.2pt/1.2pt]\addlinespace[2pt]
Student & 29.7 & 12.8 & 40.8 & 15.3 & 15.1 \\
DP-DMD & 27.5 & 12.5 & 36.9 & 14.5 & 13.9 \\
\addlinespace[2pt]
\rowcolor{OursGreen} \ourstest{} & 30.9 & 13.3 & 42.5 & 15.8 & 15.5 \\
\rowcolor{OursGreen} \ourstrain{} & \underline{35.5} & \underline{15.2} & \underline{47.5} & \underline{18.3} & \underline{19.8} \\
\rowcolor{OursGreen} \ours{} & \textbf{36.3} & \textbf{15.5} & \textbf{49.0} & \textbf{18.8} & \textbf{20.1} \\
\midrule
\rowcolor{BlockGray}
\multicolumn{6}{@{}l}{{\scriptsize\bfseries\itshape SD3}} \\
\tg{Teacher} & \tg{32.4} & \tg{13.8} & \tg{43.7} & \tg{17.7} & \tg{17.3} \\
\tg{Noise Opt.} & \tg{34.2} & \tg{14.0} & \tg{40.5} & \tg{16.0} & \tg{13.8} \\
\addlinespace[1pt]\cdashline{1-6}[1.2pt/1.2pt]\addlinespace[2pt]
Student & 22.6 & 11.2 & 31.2 & 13.3 & 9.9 \\
DP-DMD & \underline{27.3} & 12.3 & \underline{39.3} & 15.4 & \underline{14.0} \\
\addlinespace[2pt]
\rowcolor{OursGreen} \ourstest{} & 26.9 & \underline{12.8} & 37.9 & \underline{15.8} & 11.8 \\
\rowcolor{OursGreen} \ourstrain{} & 27.0 & 12.2 & 36.6 & 14.7 & 11.2 \\
\rowcolor{OursGreen} \ours{} & \textbf{32.6} & \textbf{14.7} & \textbf{46.1} & \textbf{18.1} & \textbf{14.9} \\
\midrule
\rowcolor{BlockGray}
\multicolumn{6}{@{}l}{{\scriptsize\bfseries\itshape Wan 2.1}} \\
\tg{Teacher} & \tg{45.9} & \tg{15.5} & \tg{54.8} & \tg{17.2} & \tg{17.3} \\
\addlinespace[1pt]\cdashline{1-6}[1.2pt/1.2pt]\addlinespace[2pt]
Student & 24.3 & 9.8 & 35.2 & 12.0 & 10.9 \\
DFD & 22.5 & 9.6 & 33.9 & 10.1 & 8.8 \\
\addlinespace[2pt]
\rowcolor{OursGreen} \ourstest{} & 27.0 & 11.0 & 37.9 & 13.0 & 11.0 \\
\rowcolor{OursGreen} \ourstrain{} & \underline{40.3} & \underline{15.1} & \underline{51.3} & \underline{17.4} & \underline{17.1} \\
\rowcolor{OursGreen} \ours{} & \textbf{43.9} & \textbf{15.5} & \textbf{54.6} & \textbf{17.9} & \textbf{17.6} \\
\midrule
\rowcolor{BlockGray}
\multicolumn{6}{@{}l}{{\scriptsize\bfseries\itshape MiniMax-H3}} \\
\tg{Teacher} & \tg{38.9} & \tg{12.8} & \tg{50.8} & \tg{18.3} & \tg{15.2} \\
\addlinespace[1pt]\cdashline{1-6}[1.2pt/1.2pt]\addlinespace[2pt]
Student & 24.7 & 10.0 & 39.0 & 16.8 & 12.0 \\
DFD & 24.4 & 9.7 & 37.3 & 16.1 & 11.7 \\
\addlinespace[2pt]
\rowcolor{OursGreen} \ourstest{} & 28.2 & 11.2 & 43.7 & \underline{18.5} & 12.5 \\
\rowcolor{OursGreen} \ourstrain{} & \underline{32.9} & \underline{11.7} & \underline{46.7} & 17.5 & \underline{13.5} \\
\rowcolor{OursGreen} \ours{} & \textbf{36.2} & \textbf{12.5} & \textbf{51.3} & \textbf{18.8} & \textbf{14.4} \\
\bottomrule
\end{tabularx}
\end{table}

\section{Additional Qualitative Results}
\label{app:qualitative}

\subsection{Video}
\label{app:qual_video}
\begin{figure}[H]
  \centering
  \includegraphics[width=\linewidth]{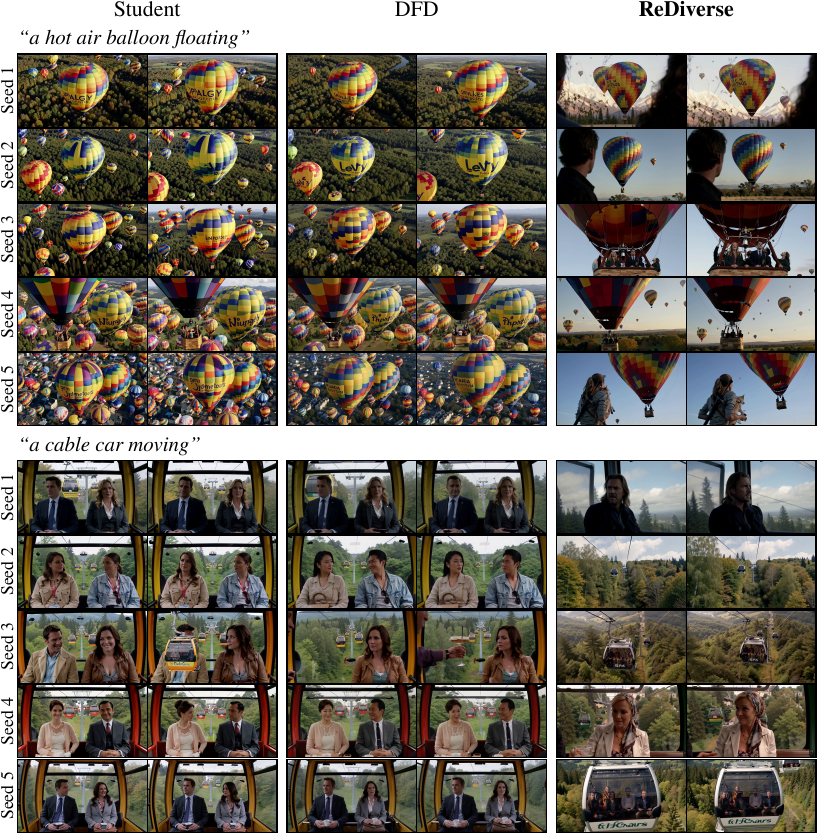}
  \caption{\textbf{Qualitative results on MiniMax-H3 (1/2).} Same protocol
    as Fig.~\ref{fig:qualitative}: Rows are seeds, each clip shown by its first and last
    frame, same noise for all three columns. Best viewed as videos available in supplementary material.}
  \label{fig:qualitative_h3_more}
\end{figure}
\begin{figure}[H]
  \centering
  \includegraphics[width=\linewidth]{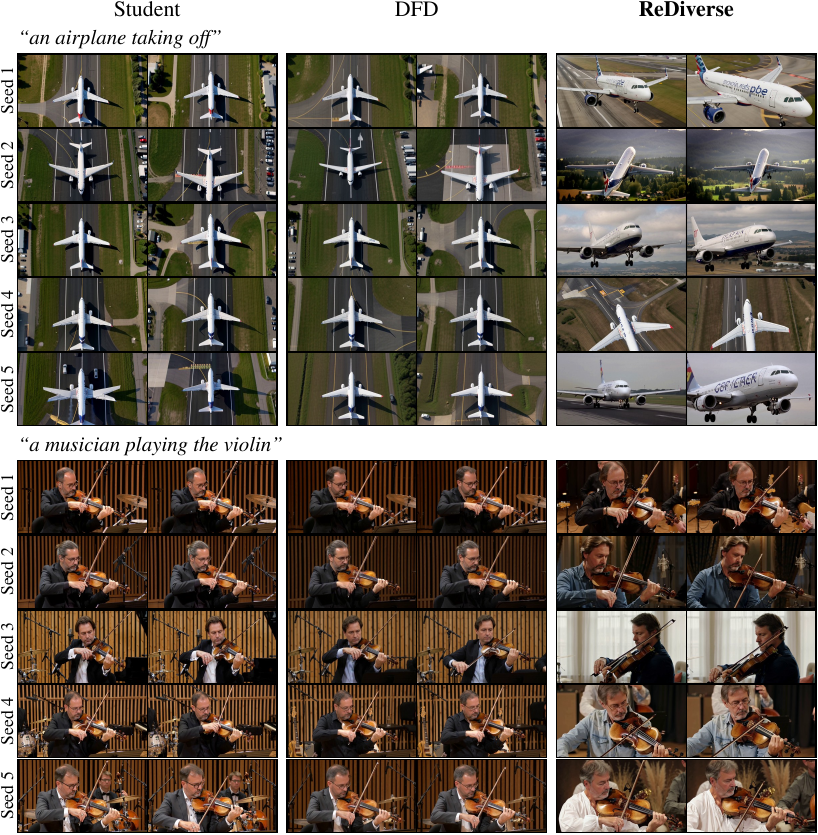}
  \caption{\textbf{Qualitative results on MiniMax-H3 (2/2).} Same protocol
    as Fig.~\ref{fig:qualitative_h3_more}. Rows are seeds, each clip shown by its first and last
    frame, same noise for all three columns. Best viewed as videos available in supplementary material.}
  \label{fig:qualitative_h3_more_b}
\end{figure}

\begin{figure}[H]
  \centering
  \includegraphics[width=\linewidth]{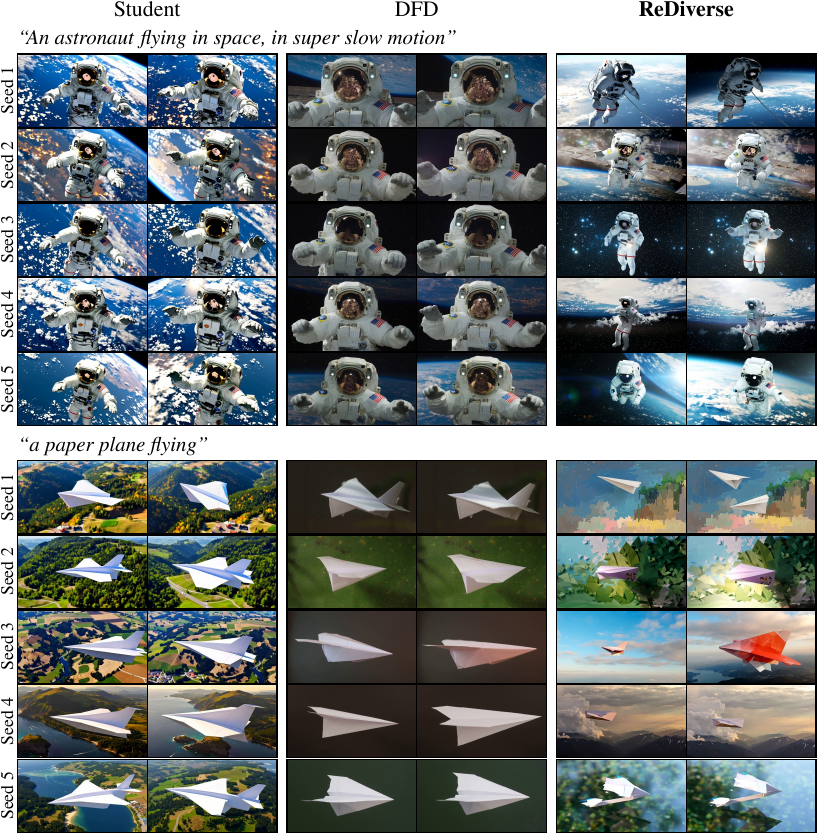}
  \caption{\textbf{Qualitative results on Wan~2.1 (1/2).} Rows are seeds, each clip shown by its first and last
    frame, same noise for all three columns. Best viewed as videos available in supplementary material.}
  \label{fig:qualitative_wan}
\end{figure}

\begin{figure}[H]
  \centering
  \includegraphics[width=\linewidth]{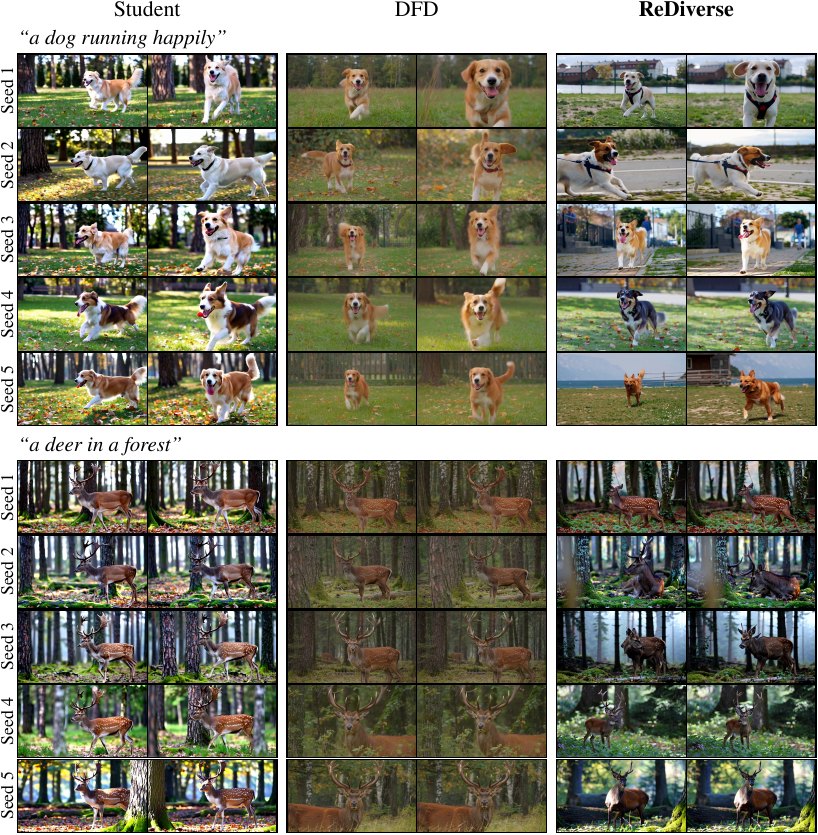}
  \caption{\textbf{Qualitative results on Wan~2.1 (2/2).} Rows are seeds, each clip shown by its first and last
    frame, same noise for all three columns. Best viewed as videos available in supplementary material.}
  \label{fig:qualitative_wan_b}
\end{figure}

\subsection{Images}
\label{app:qual_images}
\begin{figure}[H]
  \centering
  \includegraphics[width=0.92\linewidth]{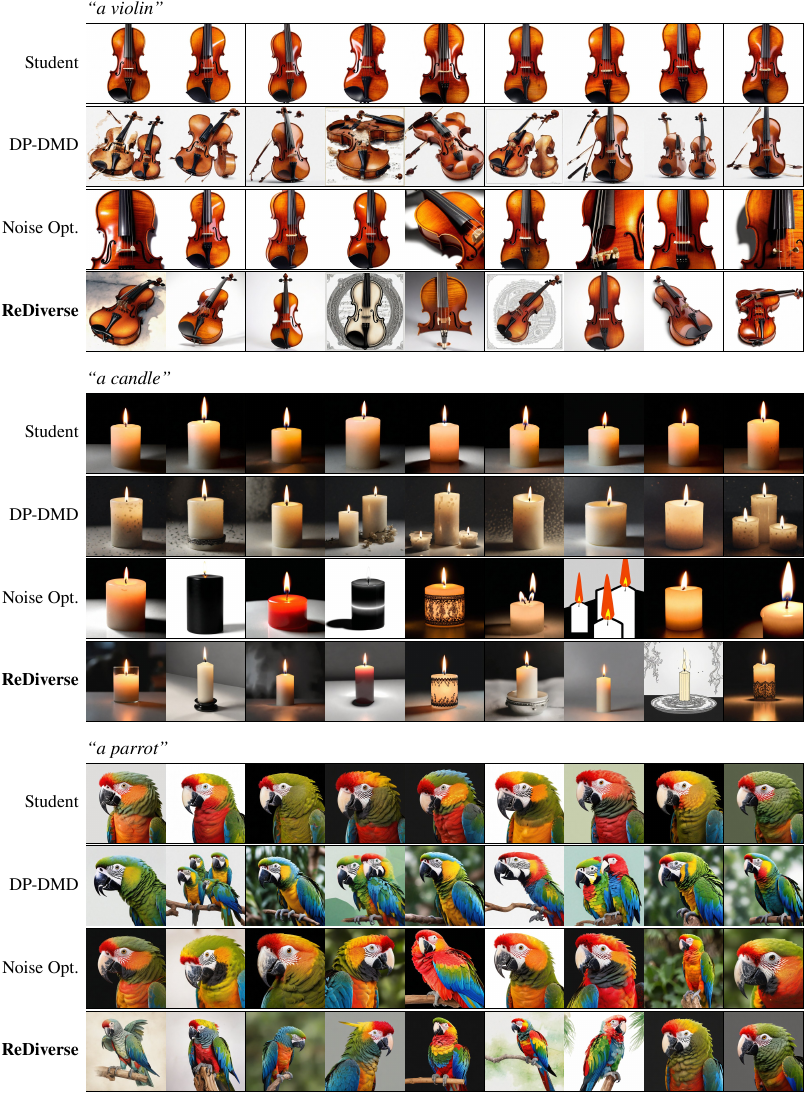}
  \caption{\textbf{Qualitative results on SDXL.} Three prompts and four 4-step
    methods are shown across nine columns. Each column corresponds to one
    initial noise, shared across all four methods.}
  \label{fig:qualitative_sdxl}
\end{figure}
\begin{figure}[H]
  \centering
  \includegraphics[width=0.92\linewidth]{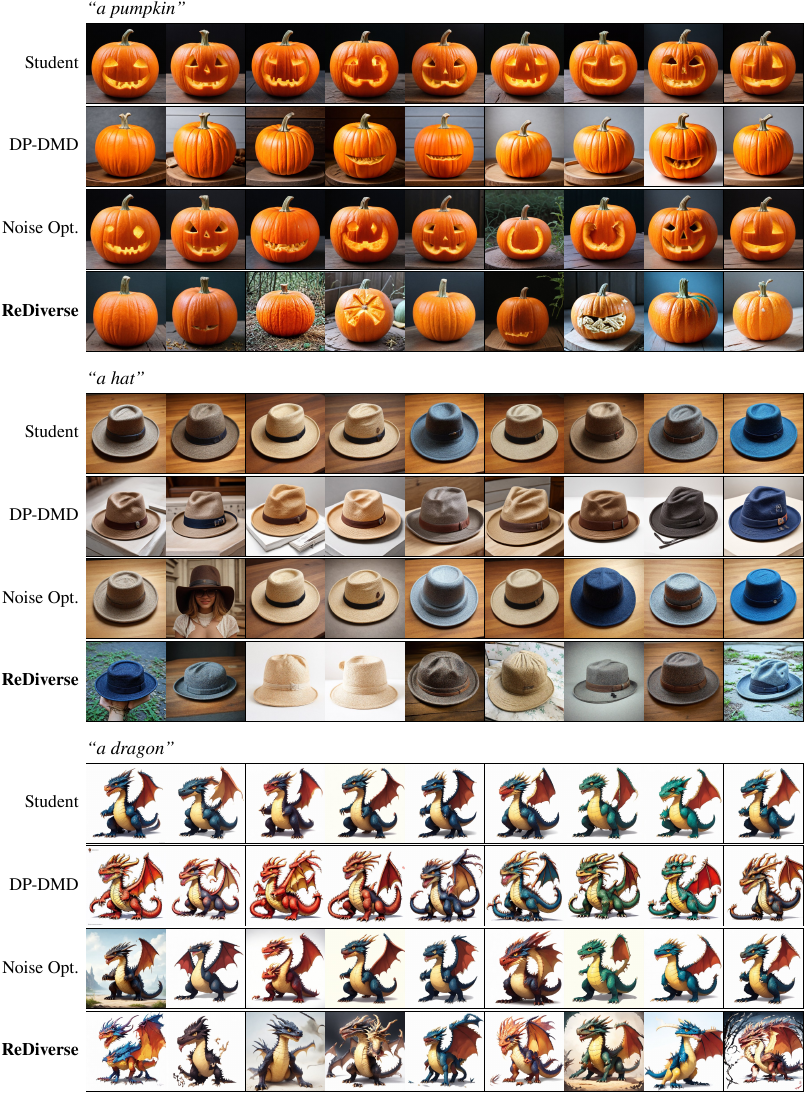}
  \caption{\textbf{Qualitative results on SD3.} Three prompts and four 4-step
    methods are shown across nine columns. Each column corresponds to one
    initial noise, shared across all four methods.}
  \label{fig:qualitative_sd3}
\end{figure}

\clearpage
\section{User Study Details}
\label{app:user_study}

\begin{figure}[t]
  \centering
  \includegraphics[height=1.3in]{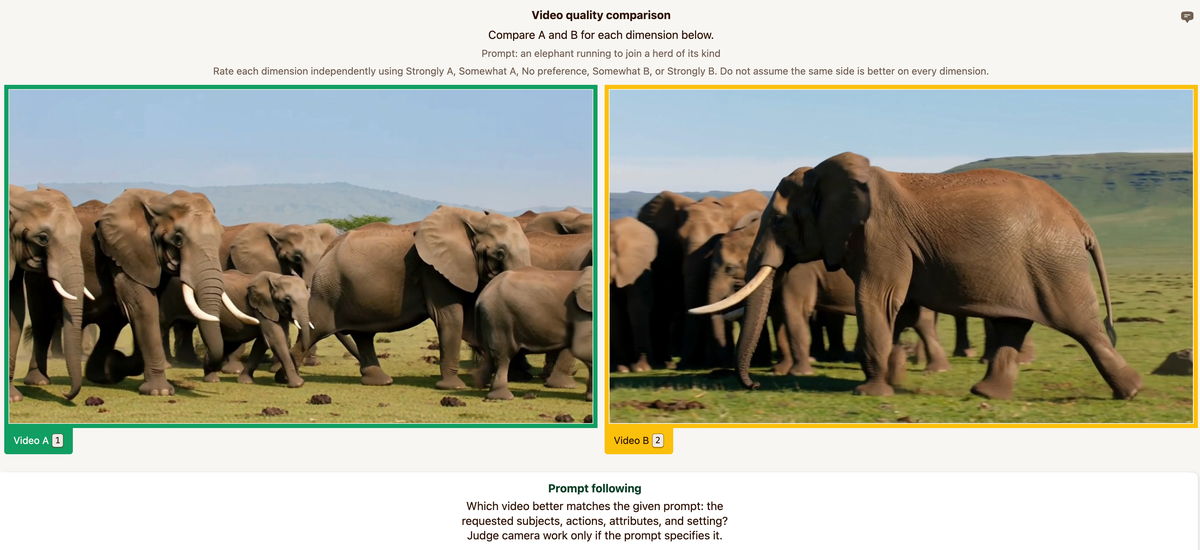}\hfill
  \includegraphics[height=1.3in]{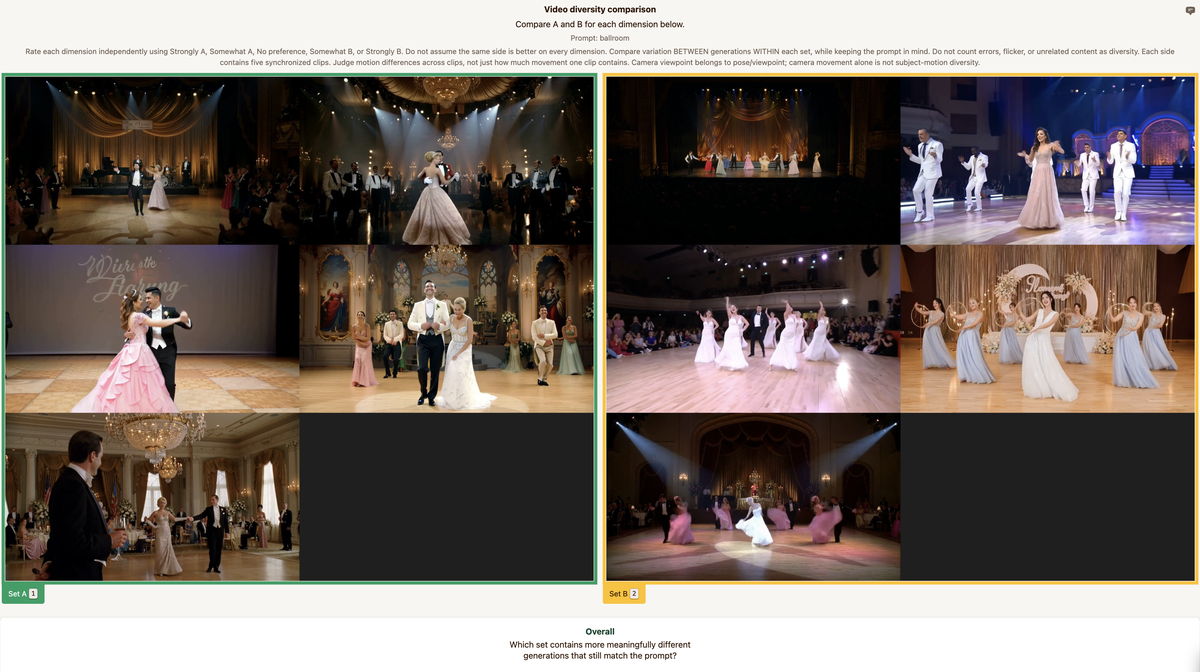}
  \caption{\textbf{The rating interface.} Left: quality comparison, with one sample per side at a shared seed.
  Right: diversity comparison, with five synchronized clips per side at the same five seeds.
  Each dimension is rated independently by choosing A, B, or No Preference.
  The instructions also specify what does not count as diversity.}
  \label{fig:study_ui}
\end{figure}

Automatic diversity metrics are only proxies and are not designed to capture motion, so we additionally evaluated diversity with human raters. Each trial compares \method{} with one baseline, the teacher, the student, or a few-step method (DP-DMD \citep{wu2026diversitypreserved} or Noise Optimization \citep{harrington2026s} for images; DFD \citep{chen2026dataforcing} for videos), using one prompt and the same initial noises. We use ten prompts per model: generic single-object GenEval prompts for images and VBench scenes/actions for videos, chosen to leave room for diversity.

\paragraph{Instruments.}
Quality and diversity were collected as separate rating jobs (Fig.~\ref{fig:study_ui}).
The quality job asks about realism and prompt following. The diversity job asks about overall diversity,
subject, pose and viewpoint, background, and, for videos, motion. The rubric explicitly excludes errors,
flicker, and unrelated content from diversity.

\paragraph{Raters and aggregation.}
Each job was rated by 30 or 31 raters, 59 distinct people in total.
The A/B order was randomized. ``No preference''
counts as half, so 50\% represents an even split.        %
\fi
\end{document}